%% file: main.tex
\documentclass[]{style/style}
\microtypesetup{expansion=false}

\usepackage{amsmath}
\usepackage{amssymb}
\usepackage{mathtools}
\usepackage{amsthm}
\usepackage{xspace}
\usepackage{colortbl}
\usepackage{multirow}
\usepackage{enumitem}
\usepackage{wrapfig}
\usepackage{array}
\usepackage{tabularx}
\usepackage{overpic}
\usepackage{bbding}

\tcbuselibrary{most}
\usepackage{algorithm}
\usepackage{algpseudocode}
\algrenewcommand\algorithmicrequire{\textbf{Input:}}
\algrenewcommand\algorithmicensure{\textbf{Output:}}
\usepackage{xspace}
\newcommand{\eg}{\textit{e.g.},\xspace}
\usepackage[table]{xcolor} 
\usepackage{makecell}      
\usepackage{tcolorbox}
\tcbuselibrary{listings}
\usepackage{graphicx}
\usepackage{booktabs}
\usepackage{pifont}
\usepackage{color}
\usepackage{arydshln}
\usepackage{makecell}
\usepackage{times}
\usepackage{epsfig}
\usepackage[T1]{fontenc}
\usepackage{hhline}
\usepackage{xcolor}
\usepackage{pifont}
\usepackage{colortbl}
\usepackage{verbatim}
\usepackage{bm} 
\usepackage{hyperref}
\usepackage{url}
\usepackage{wrapfig}
\definecolor{mygray}{gray}{.92}
\usepackage{mathtools}
\input{math_commands.tex}
\newcommand{\cmark}{\textcolor{green!55!black}{\ding{51}}}
\newcommand{\xmark}{\textcolor{red!75!black}{\ding{55}}}

\newcommand{\bench}{PEBench}

\graphicspath{{figures/}}
\makeatletter
\newcommand{\authornl}[2][]{%
  \begingroup
    \let\protect\@unexpandable@protect
    \xdef\authorlist{\authorlist\protect\\ \authorformat[#1]{#2}}%
  \endgroup
}
\newcommand{\affiliationnl}[2][]{%
  \begingroup
    \let\protect\@unexpandable@protect
    \xdef\affiliationlist{\affiliationlist\protect\\ \affiliationformat[#1]{#2}}%
  \endgroup
}
\makeatother

\title{Towards Unified Evaluation of Prompt Enhancers for Video Generation}

\author[1,2*]{Yawen~Shao}
\author[2,3*]{Yubo~Zhu}
\author[2,4*]{Ziyun~Dai}
\author[1,2*]{Zixun~Fang}
\author[2\dagger\ddagger]{Kai~Zhu}
\authornl[2]{Zeyinzi~Jiang}
\author[2]{Yufeng~Ai}
\author[2]{Siyang~Sun}
\author[2]{Haolan~Xue}
\author[2]{Yu~Shang}
\author[2]{Yuxiang~Bao}
\authornl[2]{Zoubin~Bi}
\author[2]{Jingming~Luo}
\author[2]{Jie~Xiao}
\author[2]{Chaojie~Mao}
\author[5]{Zhehan~Kan}
\author[6]{Hongchen~Luo}
\authornl[2]{Yu~Liu}
\author[3]{Sheng~Zhong}
\author[3\ddagger]{Wei~Tong}
\author[1]{Xueyang~Fu}
\author[1]{Yang~Cao}
\author[1\ddagger]{Wei~Zhai}
\author[1]{Zheng-Jun~Zha}

\affiliation[1]{University of Science and Technology of China}
\affiliation[2]{Wan Team, Alibaba Group}
\affiliationnl[3]{Nanjing University}
\affiliation[4]{Fudan University}
\affiliation[5]{Tsinghua University}
\affiliation[6]{Northeastern University}

\renewcommand\contribution[2][]{\addtolist[#1]{#2}{\contributionlist}{\contributionformat}{,\quad}}
\contribution[*]{Equal contribution}
\contribution[\dagger]{Project leader}
\contribution[\ddagger]{Corresponding author}

\abstract{
Modern video generators can realize increasingly complex visual narratives, positioning the prompt enhancer (PE) as a critical bridge from concise user instructions and multimodal references to structured cinematic plans. However, existing PE evaluation relies on rendered videos, imposing substantial computational and human costs, slowing PE training and iteration, and conflating PE quality with downstream generator behavior. To address this gap, we introduce \bench{}, the first unified benchmark for direct PE evaluation across text-to-video, image-to-video, and reference-to-video prompt enhancement. It comprises 1,100 expert-verified cases and 1,005 visual assets, spanning 35 fine-grained tasks with diverse temporal, cinematic, audiovisual, and multi-reference requirements. In addition, we develop \bench{} evaluation, an evidence-grounded framework that combines modality-aware fact extraction with rubric-based assessment across 24 criteria. Our systematic evaluation of representative open- and closed-source PE methods reveals an emerging shift from fine-grained descriptive expansion toward intent-preserving cinematic planning, while the caption-reconstruction and forward-refinement methods show complementary strengths in cinematic coverage and semantic fidelity or internal coherence, respectively. Human validation shows that \bench{} scores align closely with expert judgments of enhanced prompts and downstream videos from Wan3.0 and MiniMax-H3, indicating that prompt-level evaluation reliably reflects downstream utility.

Project page: \url{https://github.com/yawen-shao/PEBench}
}

\begin{document}

\maketitle
\makeatletter
\makeatother

\input{section/introduction.tex}
\input{section/related_work.tex}
\input{section/methodology.tex}
\input{section/experiments.tex}
\input{section/conclusion.tex}

\clearpage
\bibliographystyle{unsrtnat}
\bibliography{paper}

\clearpage
\input{section/appendix.tex}

\end{document}

%% file: math_commands.tex
\usepackage{amsmath,amsfonts,bm}

\def\eqref#1{equation~\ref{#1}}

\def\1{\bm{1}}

\DeclareMathAlphabet{\mathsfit}{\encodingdefault}{\sfdefault}{m}{sl}
\SetMathAlphabet{\mathsfit}{bold}{\encodingdefault}{\sfdefault}{bx}{n}

%% file: section/introduction.tex
\section{Introduction}
Rapid advances in large-scale video generation models~\citep{wan2026,bytedance2026seedance25,minimaxh3} have transformed content creation by enabling diverse video synthesis from natural-language instructions and multimodal references.
Despite their remarkable capabilities, the quality and fidelity of generated videos remain strongly influenced by input conditions that adequately express user intent and generation requirements~\citep{hunyuan,zyb_wanpe_instead}.
Concise or ambiguous user instructions can lead generators to misinterpret the bindings between subjects and actions, the intended event order, or the roles of visual references, revealing a mismatch between user intent and model interpretation.
A prompt enhancer (PE) addresses this mismatch by converting user instructions and conditioning inputs into an enhanced prompt that guides downstream generation, making PE quality an important factor in high-quality video synthesis~\citep{zyb_wanpe_instead}.

However, PE development remains constrained by an indirect evaluation loop: validating each update requires downstream video synthesis followed by automatic scoring or blind human comparisons~\citep{zyb_wanpe_instead,vpo,scmapr,rapo,promptavideo,rapo++}. These computational and human costs delay feedback for training and refinement.
Beyond this inefficiency, rendered videos reflect the joint influence of PE quality and generator capability, making observed errors difficult to attribute reliably to either component.
These limitations motivate direct evaluation of enhanced prompts to assess PE quality independently of generator behavior and provide rapid, PE-specific feedback throughout iterative refinement.

\begin{figure*}[t]
\centering  
    \begin{overpic}[width=1\linewidth]{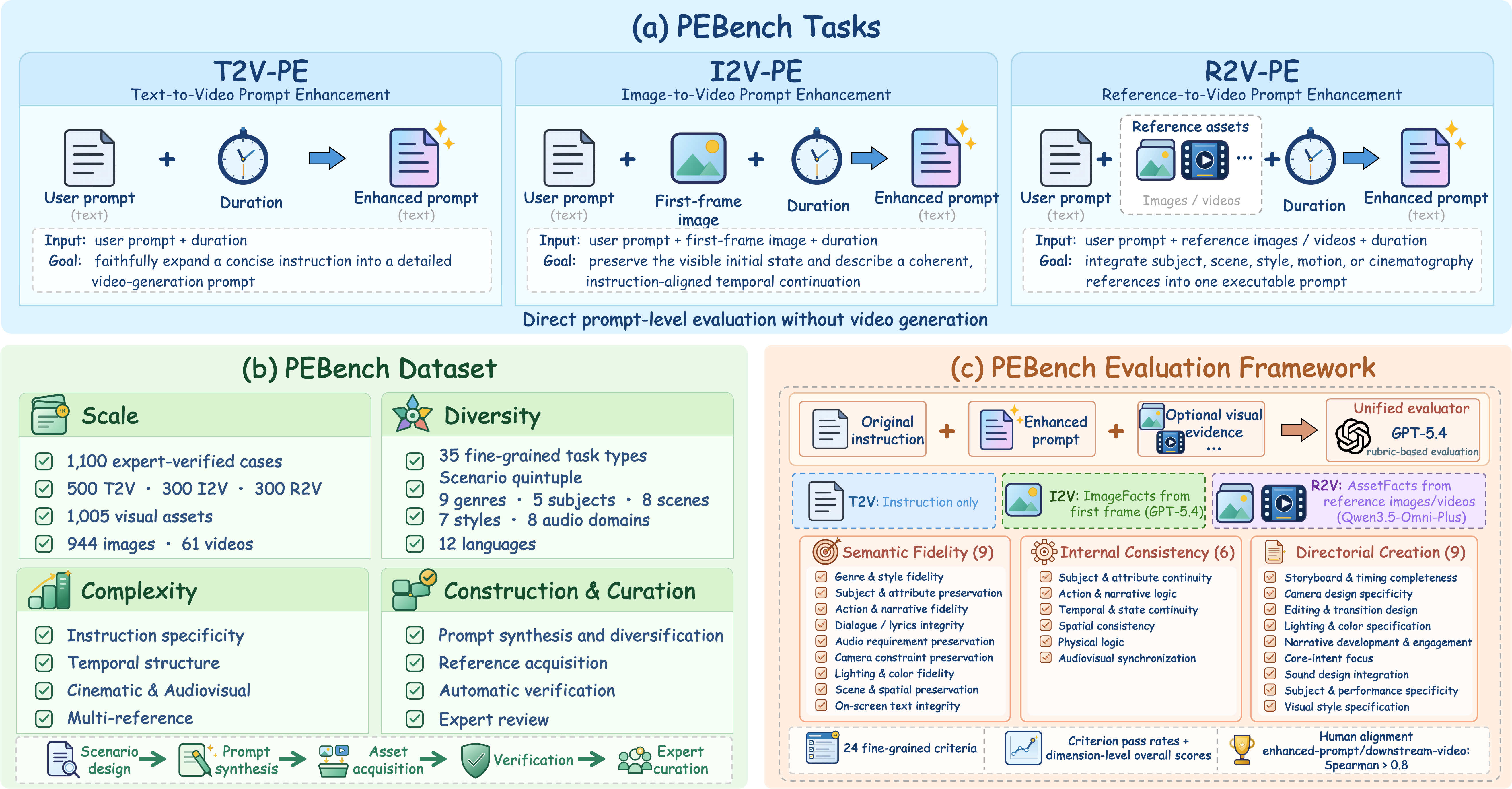}
    \end{overpic} 
    \caption{\textbf{Overview of \bench{}.} (a) \bench{} unifies T2V-PE, I2V-PE, and R2V-PE for direct prompt-level evaluation without video generation. (b) The dataset contains 1,100 expert-verified cases and 1,005 visual assets, covering diverse tasks, content, and conditioning complexity through a four-stage construction pipeline. (c) The evidence-grounded framework combines modality-aware evidence construction with rubric-based evaluation across 24 fine-grained criteria organized into Semantic Fidelity, Internal Consistency, and Directorial Creation.}
    \label{fig:overview}
\end{figure*}

In this paper, we introduce \textbf{\bench{}}, the first unified benchmark for direct evaluation of video prompt enhancers.
As illustrated in Figure~\ref{fig:overview}(a), \bench{} spans T2V-PE, I2V-PE, and R2V-PE, covering text-only, first-frame, and multi-reference conditioning.
Figure~\ref{fig:overview}(b) summarizes 1,100 expert-verified cases and 1,005 visual assets, covering 35 fine-grained task types with broad variation in content, language, and conditioning complexity. 
To ensure diversity and quality, we employ a staged dataset construction pipeline combining broad scenario coverage with rigorous verification.
Beyond dataset construction, the evidence-grounded framework in Figure~\ref{fig:overview}(c) evaluates each enhanced prompt against the original instruction and modality-aware facts extracted from visual inputs where applicable.
Its 24 fine-grained criteria form three complementary dimensions: Semantic Fidelity verifies the preservation of source requirements, Internal Consistency examines narrative, spatiotemporal, and audiovisual coherence, and Directorial Creation assesses the quality of creative guidance for realizing the intended video. A unified rubric-based evaluator reports criterion pass rates and dimension scores, yielding an interpretable profile of PE quality.

Using \bench{}, we systematically evaluate representative open- and closed-source PE systems across T2V-PE, I2V-PE, and R2V-PE, revealing four key findings, most notably an emerging shift toward cinematic planning among industry-developed PE and observed links between specific training strategies and capability profiles.
(i) Prompt enhancement is evolving from fine-grained descriptive expansion toward intent-preserving cinematic planning as video generators become increasingly capable.
(ii) Caption-reconstruction methods achieve broader cinematic coverage, whereas forward-refinement methods favor semantic preservation or internal coherence.
(iii) Increasing temporal and reference complexity generally degrades PE performance.
(iv) \bench{} evaluation scores achieve Spearman correlations above 0.8 with expert rankings of enhanced prompts and downstream videos produced by Wan3.0 and MiniMax-H3, indicating that prompt-level evaluation reliably reflects downstream utility across generators.

Our contributions are summarized as follows:

\begin{itemize}
     \item We introduce \bench{}, the first unified benchmark for direct evaluation of video prompt enhancers across T2V-PE, I2V-PE, and R2V-PE, comprising 1,100 expert-verified cases and 1,005 visual conditioning assets spanning 35 fine-grained task types.
    \item We propose \bench{} Evaluation, an evidence-grounded framework that combines modality-aware fact extraction for visually conditioned tasks with rubric-based assessment.
    \item We systematically evaluate representative PE, revealing a shift toward cinematic planning and capability differences associated with training strategies, while confirming human alignment and downstream utility across generators.
\end{itemize}

%% file: section/related_work.tex
\section{Related work}
\textbf{Prompt Enhancement for Video Generation.}
As video generators increasingly support temporally structured and multimodally conditioned content~\citep{wan2026,bytedance2026seedance25,klingai2026}, prompt enhancers (PE) have become important for translating concise user intent into detailed instructions for downstream video generation.
Existing approaches~\citep{VidPrompter,CAPE-T2V,long2025vistatesttimeselfimprovingvideo} broadly follow training-based or inference-time paradigms.
Training-based methods learn prompt rewriting from constructed supervision or preference signals.
VPO~\citep{vpo} and Prompt-A-Video~\citep{promptavideo} optimize LLM-based rewriters using supervised and preference objectives.
WanPE~\citep{zyb_wanpe_instead} instead reconstructs detailed video captions from compressed prompts using video-grounded inverse pairs, while LingBot-Video-PE~\citep{lingbotvideo} maps expanded prompts into model-specific structured captions.
In contrast, inference-time methods adopt diverse strategies:
RAPO~\citep{rapo} retrieves relevant examples, while SCMaPR~\citep{scmapr} performs agentic verification and correction. LTX-2.5-PE~\citep{ltx25} directly rewrites prompts using its integrated Gemma 4 text encoder~\citep{gemmateam2026gemma4technicalreport}, whereas H3-Context-IR~\citep{minimaxh3} employs a multi-stage multimodal workflow. 


\textbf{Benchmarks for Video Generation and Prompt Enhancement.}
Video-generation benchmarks have evolved from assessing T2V quality and prompt alignment~\citep{vbench,liu2024evalcrafterbenchmarkingevaluatinglarge,han2025videobenchhumanalignedvideogeneration} to covering reference-conditioned, multi-shot, audio-video, and streaming generation~\citep{yuan2025opens2vnexusdetailedbenchmarkmillionscale,wei2026univbenchunifiedevaluationvideo,wei2026msavbenchcomprehensivereliableevaluation,liu2026streamavbenchcomprehensivebenchmarkstreaming}.
However, these benchmarks are designed to assess rendered videos rather than enhanced prompts.
Applying them to PE evaluation requires an additional generation stage, increasing evaluation cost and latency while entangling PE performance with downstream generator behavior.
As summarized in Table~\ref{tab:benchmark_comparison}, \bench{} instead evaluates enhanced prompts directly across T2V-PE, I2V-PE, and R2V-PE using standardized rubrics covering 24 criteria, enabling fine-grained assessment independent of downstream video generation while offering broad coverage of prompt structures and conditioning modalities at scale.

\newcommand{\singlehead}[1]{\raisebox{0.7ex}{\textbf{#1}}}
\begin{table*}[t]
\centering
\caption{
\textbf{Comparison with existing video-generation benchmarks.}
\textbf{Cine.}: cinematic and camera control;
\textbf{Scripted}: temporal or shot-level prompts;
\textbf{Audio Req.}: explicit audio requirements in prompts;
\textbf{Multi-lang.}: three or more languages;
\textbf{Video Ref.}: conditioning on reference videos;
\textbf{Direct PE}: direct prompt-level PE evaluation decoupled from downstream video generators.
}
\label{tab:benchmark_comparison}

\vspace{0.4em}
\resizebox{\linewidth}{!}{%
\begin{tabular}{
l!{\color{black}\vrule width 0.4pt}
*{10}{c}
!{\color{black}\vrule width 0.4pt}
ccc
}
\toprule
\singlehead{Benchmark}
& \singlehead{T2V}
& \singlehead{I2V}
& \singlehead{R2V}
& \singlehead{Cine.}
& \singlehead{Scripted}
& \shortstack{\textbf{Multi-}\\\textbf{shot}}
& \shortstack{\textbf{Audio}\\\textbf{Req.}}
& \shortstack{\textbf{Multi-}\\\textbf{lang.}}
& \shortstack{\textbf{Video}\\\textbf{Ref.}}
& \shortstack{\textbf{Direct}\\\textbf{PE}}
& \shortstack{\textbf{\#}\\\textbf{Metrics}}
& \shortstack{\textbf{\#}\\\textbf{Prompts}}
& \shortstack{\textbf{\# Ref.}\\\textbf{Assets}} \\
\midrule

VBench~\citep{vbench}
& \cmark & \xmark & \xmark
& \xmark & \xmark & \xmark & \xmark
& \xmark & \xmark & \xmark
& 16 & $\sim$1,600 & -- \\

EvalCrafter~\citep{liu2024evalcrafterbenchmarkingevaluatinglarge}
& \cmark & \xmark & \xmark
& \xmark & \xmark & \xmark & \xmark
& \xmark & \xmark & \xmark
& 17 & 700 & -- \\

Video-Bench~\citep{han2025videobenchhumanalignedvideogeneration}
& \cmark & \xmark & \xmark
& \xmark & \xmark & \xmark & \xmark
& \xmark & \xmark & \xmark
& 9 & 419 & -- \\

OpenS2V-Nexus~\citep{yuan2025opens2vnexusdetailedbenchmarkmillionscale}
& \cmark & \xmark & \cmark
& \xmark & \xmark & \xmark & \xmark
& \xmark & \xmark & \xmark
& 6 & 180 & 180 \\

UniVBench~\citep{wei2026univbenchunifiedevaluationvideo}
& \cmark & \xmark & \cmark
& \cmark & \cmark & \cmark & \xmark
& \xmark & \xmark & \xmark
& 21 & 200 & 864 \\

MSAVBench~\citep{wei2026msavbenchcomprehensivereliableevaluation}
& \cmark & \xmark & \cmark
& \cmark & \cmark & \cmark & \cmark
& \cmark & \xmark & \xmark
& 20 & 286 & 165 \\

StreamAV-Bench~\citep{liu2026streamavbenchcomprehensivebenchmarkstreaming}
& \cmark & \xmark & \xmark
& \xmark & \cmark & \xmark & \cmark
& \xmark & \xmark & \xmark
& 32 & 320 & -- \\

\midrule
\rowcolor{gray!12}
\textbf{\bench{} (Ours)}
& \cmark
& \cmark
& \cmark
& \cmark
& \cmark
& \cmark
& \cmark
& \cmark
& \cmark
& \cmark
& \textbf{24}
& \textbf{1,100}
& \textbf{1,005} \\

\bottomrule
\end{tabular}%
}

\vspace{0.3em}
\end{table*}

%% file: section/methodology.tex
\section{\bench{}}
\subsection{Task Definition}
\label{sec:task_definition}
A video prompt enhancer takes a user prompt, optional reference assets $\mathcal{R}$, and a target duration as input and produces an enhanced prompt for a downstream video generator, as illustrated in Figure~\ref{fig:overview}(a).
\bench{} evaluates this output directly without requiring video generation, thereby separating PE quality from downstream generator behavior. Specifically, it assesses whether the enhanced prompt preserves user intent and conditioning constraints, remains internally consistent, and provides sufficient detail to specify the intended video.

To provide comprehensive coverage, \bench{} introduces a unified benchmark for evaluating prompt enhancers across three distinct task settings:

\begin{itemize}
    \item \textbf{Text-to-Video Prompt Enhancement (T2V-PE).}
    Given a user prompt and a target duration, with no reference assets ($\mathcal{R}=\varnothing$), the PE should produce a detailed prompt that faithfully expands the requested video content.

    \item \textbf{Image-to-Video Prompt Enhancement (I2V-PE).}
    Given a user prompt, a first-frame image ($\mathcal{R}=\{I_0\}$), and a target duration, the PE should preserve the visible initial state and describe a coherent, instruction-aligned temporal continuation.

    \item \textbf{Reference-to-Video Prompt Enhancement (R2V-PE).}
    Given a user prompt, a set of reference images or videos ($\mathcal{R}=\{a_i\}_{i=1}^{N}$), and a target duration, the PE should incorporate each asset according to its specified role, such as subject, scene, style, motion, or cinematography, into a single coherent generation prompt.
\end{itemize}

\begin{figure*}[t]
\centering  
    \begin{overpic}[width=1\linewidth]{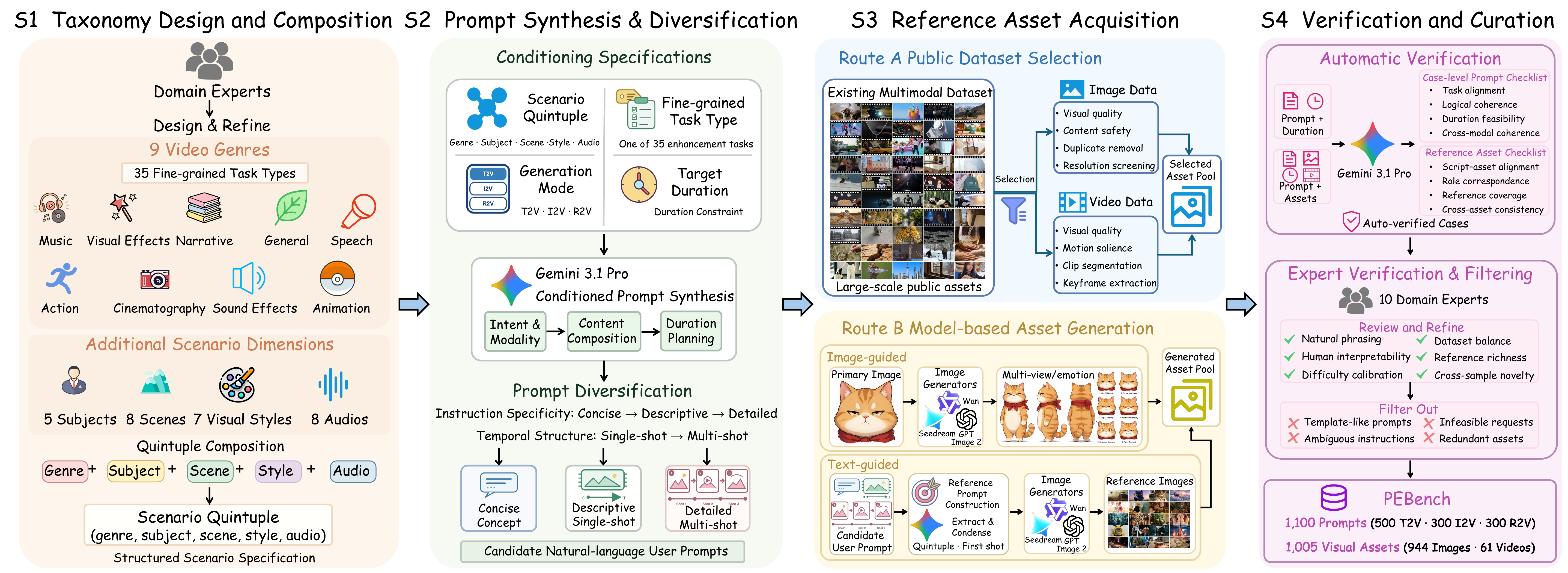}
    \end{overpic} 
    \caption{\textbf{\bench{} data construction pipeline.} The four-stage pipeline comprises taxonomy design and composition, prompt synthesis and diversification, reference asset acquisition from public datasets and generative models, and automated verification followed by expert curation.}
    \label{fig:data_construct}
\end{figure*}

\subsection{Data Construction}
To construct a diverse and reliable benchmark, we develop a four-stage pipeline that combines automated synthesis and verification with expert-driven design and curation, as illustrated in Figure~\ref{fig:data_construct}.

\textbf{Stage 1: Taxonomy design and composition.}
Domain experts define nine video genres and refine them into 35 fine-grained task types to establish broad and structured task coverage.
They further specify five subject categories, eight scene domains, seven visual styles, and eight audio domains to guide scenario composition. Using the genre and these four additional dimensions, each scenario is represented as a $(\text{genre}, \text{subject}, \text{scene}, \text{style}, \text{audio})$ quintuple.

\textbf{Stage 2: Prompt synthesis and diversification.} We sample a scenario quintuple and associate it with a generation mode (T2V, I2V, or R2V), a fine-grained task type, and a target duration. Gemini 3.1 Pro~\citep{gemini31pro} then synthesizes a user prompt from these specifications, jointly considering intent and modality, content composition, and duration planning.
To reflect real-world prompting behavior, we vary instruction specificity from concise to detailed and temporal structure between single- and multi-shot formats, producing inputs ranging from short concepts to structured scripts.

\textbf{Stage 3: Reference asset acquisition.}
We construct a diverse reference pool through two complementary routes. Public images and videos from existing datasets~\citep{opensubject,openvid,yuan2025opens2vnexusdetailedbenchmarkmillionscale,goku} undergo modality-specific curation for quality, safety, and reference suitability. Model-generated references produced with Wan-Image~\citep{wanimage}, Seedream~\citep{seedream}, and GPT Image 2~\citep{gptimage2} follow image-guided and text-guided pipelines to create identity-consistent variants and scenario-aligned assets. The resulting pool supports single-image, multi-image, video-only, and hybrid image-video conditioning.

\textbf{Stage 4: Verification and curation.}
Each candidate undergoes a two-stage quality-control process combining scalable model-based screening with expert review. Gemini 3.1 Pro first derives case-specific checks from the instruction and target duration to detect infeasible or inconsistent specifications. For reference-conditioned cases, it jointly inspects the instruction and candidate assets to verify role correspondence, script-asset alignment, reference coverage, and cross-asset consistency. Ten domain experts then refine the verified cases for clarity and difficulty, removing ambiguous, template-like, infeasible, or redundant content while preserving dataset balance and diversity. 

\begin{figure*}[t]
\centering  
    \begin{overpic}[width=1\linewidth]{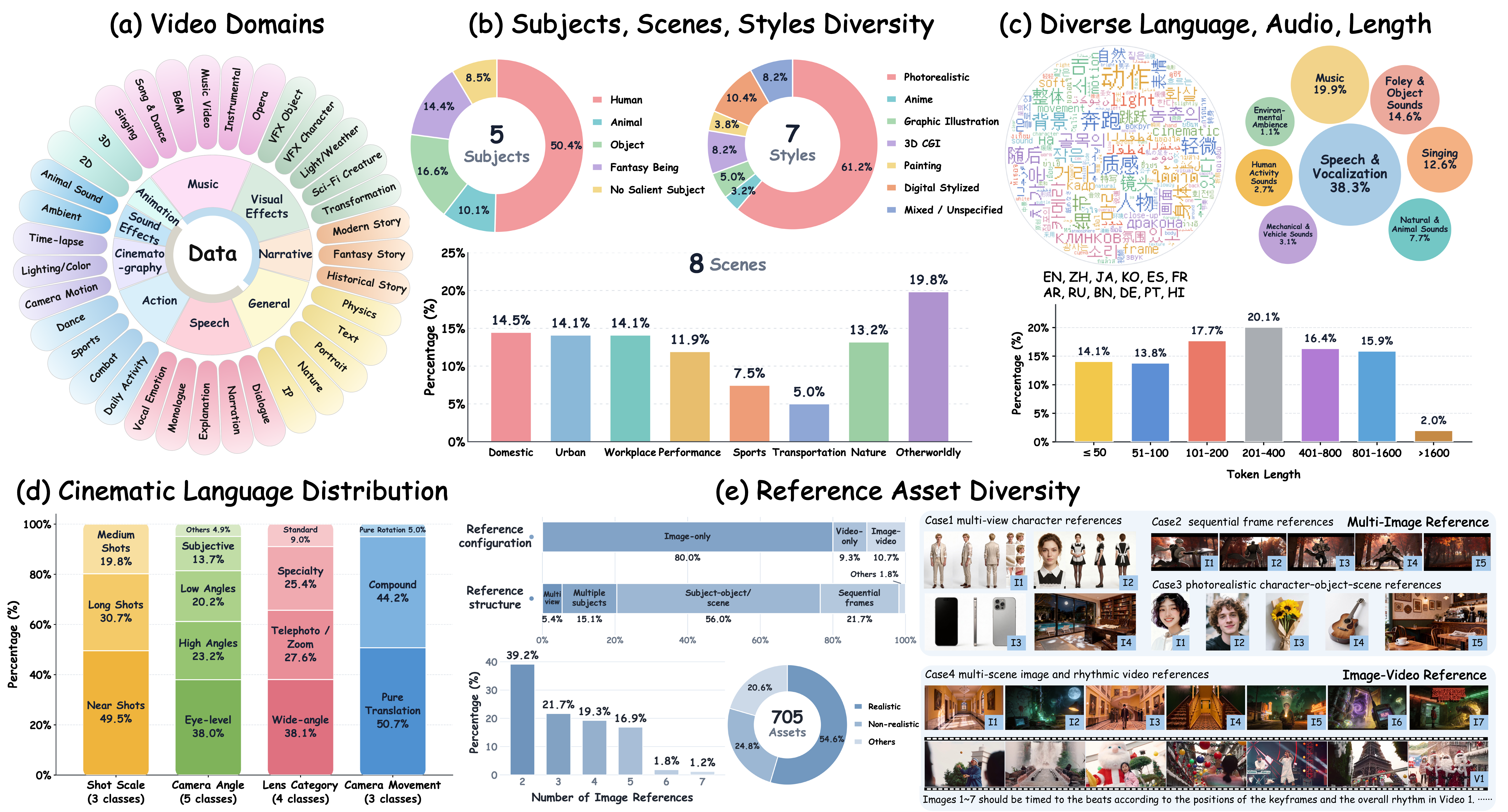}
    \end{overpic} 
    \caption{\textbf{Detailed Data Statistics and Diversity of \bench{}.} The benchmark covers a broad taxonomy of video-generation domains and fine-grained tasks (a), diverse subjects, scenes, and visual styles (b), multilingual prompts, audio conditions, and prompt lengths (c), rich cinematic requirements (d), and diverse reference assets (e). More statistics are in Appendix~\ref{app:detailed_data_analysis}.}
    \label{fig:benchmark_statistics}
\end{figure*}

\subsection{Data Statistics}
\label{sec:data_statistics}

\textbf{Scale and Diversity.}
\bench{} comprises 1,100 prompts across three PE tasks: T2V-PE (500), I2V-PE (300), and R2V-PE (300). As shown in Figure~\ref{fig:benchmark_statistics}(a,b), its taxonomy spans 9 video genres and 35 fine-grained task types, while its scenarios cover 5 subject categories (\eg human and fantasy beings), 8 scene domains (\eg domestic and nature), and 7 visual styles (\eg photorealistic and 3D CGI), providing broad task, semantic, environmental, and aesthetic diversity.

\textbf{Linguistic and Audio Diversity.}
In Figure~\ref{fig:benchmark_statistics}(c), \bench{} spans 12 languages (\eg English and Chinese) and 8 audio domains (\eg speech and vocalization, and music). Prompt lengths range from 6 to 4,120 tokens, encompassing concise instructions and long, temporally structured scripts.

\textbf{Temporal and Cinematic Complexity.}
\bench{} captures temporal and cinematic complexity through structured prompts containing 1--13 shots and rich cinematic language. As shown in Figure~\ref{fig:benchmark_statistics}(d), it covers 3 shot scales (\eg near and long shots), 5 camera angles (\eg eye-level and low angles), 4 lens types (\eg wide-angle and telephoto/zoom), and 3 camera-movement classes (\eg pure translation and compound movement).
Beyond these dimensions, the prompts also specify diverse composition patterns, transitions, lighting, and color schemes.

\textbf{Reference Asset Diversity.}
As shown in Figure~\ref{fig:benchmark_statistics}(e), \bench{} contains 1,005 visual conditioning assets, with 944 images and 61 videos. I2V-PE uses 300 first-frame images, while R2V-PE includes 644 images and 61 videos in image-only, video-only, and hybrid configurations, with up to 7 images or 2 videos per case.
Beyond modality diversity, R2V-PE covers multi-reference structures (\eg multi-view and multiple subjects) across realistic (54.6\%), non-realistic (24.8\%), and mixed or unspecified (20.6\%) styles.
Four representative cases illustrate multi-view characters, stylized frame sequences, photorealistic character–object–scene compositions, and multi-scene images paired with rhythm-guiding videos, supporting compositional, sequential, and cross-modal evaluation.

\subsection{Evaluation Framework}
\label{sec:evaluation_framework}
Evaluating prompt enhancers through generated videos conflates prompt quality with downstream generator behavior while incurring substantial computational cost and feedback latency, slowing training and iteration. \bench{} thus directly evaluates enhanced prompts along three dimensions: Semantic Fidelity (\textit{SF}), Internal Consistency (\textit{IC}), and Directorial Creation (\textit{DC}). These dimensions respectively assess fidelity to the inputs, internal coherence, and creative directorial planning for realizing the intended video. Figure~\ref{fig:overview}(c) summarizes the 24 fine-grained criteria.

\textbf{Semantic Fidelity.}
\textit{SF} assesses whether the enhanced prompt preserves the original instruction and relevant information from visual references. Genre and Style Fidelity (\textit{GSF}) measures retention of genre, medium, and visual style. Subject and Attribute Preservation (\textit{SAP}) examines subject identity, appearance, and attribute bindings, while Action and Narrative Fidelity (\textit{ANF}) covers actions, event order, and narrative intent. Dialogue and Lyrics Integrity (\textit{DLI}) verifies verbal content, language, and speaker attribution, while Audio Requirement Preservation (\textit{ARP}) covers speech, music, and sound effects. Camera Constraint Preservation (\textit{CCP}), Lighting and Color Fidelity (\textit{LCF}), Scene and Spatial Preservation (\textit{SSP}), and On-Screen Text Integrity (\textit{OTI}) respectively assess camera constraints, visual conditions, spatial relations, and visible text. In I2V-PE, the first frame anchors the initial state. In R2V-PE, each reference's instruction-specified role determines the asset facts evaluated.

\textbf{Internal Consistency.}
Whereas \textit{SF} measures fidelity to the inputs, \textit{IC} evaluates the logical and spatiotemporal coherence of the enhanced prompt.
Subject and Attribute Continuity (\textit{SAC}) assesses the stability of subject identity and attributes. Action and Narrative Logic (\textit{ANL}) evaluates event progression, causal relations, and subject interactions, while Temporal and State Continuity (\textit{TSC}) examines event order and state transitions. Spatial Consistency (\textit{SPC}) assesses locations, relative positions, and movement paths. Physical Logic (\textit{PL}) checks whether events follow real-world physics or established fictional logic. Audiovisual Synchronization (\textit{AVS}) evaluates the alignment between visual events and associated dialogue, music, and sound effects.

\textbf{Directorial Creation.}
\textit{DC} assesses whether the enhanced prompt provides relevant, explicit, and creative direction to specify the intended video across temporal, cinematic, narrative, and audiovisual aspects.
Storyboard and Timing Completeness (\textit{STC}) evaluates shot organization and duration allocation. Camera Design Specificity (\textit{CDS}) examines framing, viewpoint, focus, and camera movement, while Editing and Transition Design (\textit{ETD}) evaluates shot connections and editing logic. Lighting and Color Specification (\textit{LCS}) assesses illumination and color planning. Narrative Development and Engagement (\textit{NDE}) measures content progression and variation, while Core-Intent Focus (\textit{CIF}) checks whether added content supports the original request. Sound Design Integration (\textit{SDI}), Subject and Performance Specificity (\textit{SPS}), and Visual Style Specification (\textit{VSS}) respectively evaluate audio design, subject presentation, and visual-style articulation. 


\textbf{Evidence-Grounded Fine-Grained Evaluation.}
We formulate evaluation as a two-stage process combining modality-aware evidence construction with unified criterion-level assessment.
First, we construct this evidence using modality-specific extraction checklists. GPT-5.4~\citep{gpt54} extracts static visual facts from image-only inputs, while Qwen3.5-Omni-Plus~\citep{qwenteam2026qwen35omnitechnicalreport} jointly analyzes all assets when video references are present, capturing static, temporal, audiovisual, and cross-asset evidence. Each fact is grounded in its source asset. The resulting \textit{ImageFacts} for I2V-PE and \textit{AssetFacts} for R2V-PE are constructed once and shared across methods.
Second, GPT-5.4 serves as the unified evaluator across all three PE tasks. Given the original instruction, enhanced prompt, and applicable visual evidence, it assesses all 24 criteria using fixed criterion-specific rubrics and structured checklists. Shared evidence and fixed rubrics across methods enable consistent, interpretable criterion-level comparisons within each task setting.
%
See Appendix~\ref{app:evaluation_details} for more details.

%% file: section/experiments.tex
\section{Experiments}
\subsection{Experimental Setup}
\label{sec:experimental_setup}
We evaluate ten PE systems: H3-Context-IR~\citep{minimaxh3}, WanPE-397B and WanPE-35B~\citep{zyb_wanpe_instead}, LTX-2.5-PE~\citep{ltx25}, LingBot-Video-PE~\citep{lingbotvideo}, SCMaPR~\citep{scmapr}, VPO~\citep{vpo}, Prompt-A-Video-Cog and Prompt-A-Video-OS~\citep{promptavideo}, and RAPO~\citep{rapo}.
All systems are evaluated on T2V-PE, with multimodal-capable systems also evaluated on I2V-PE and R2V-PE.
%

\begin{table*}[t]
\centering
\scriptsize
\caption{\textbf{Main results on \bench{}.}
Fine-grained metrics report criterion-level pass rates, defined as the percentage of cases without corresponding issues.
The \textit{Overall} scores aggregate criterion-level issues according to their number and severity in each case within each dimension.
The best and second-best results for each task are highlighted in \textbf{bold} and \underline{underlined}, respectively.}
\renewcommand{\arraystretch}{1}
\setlength{\tabcolsep}{1.5pt}

\resizebox{\linewidth}{!}{%
\begin{tabular}{
l!{\color{black}\vrule width 0.4pt}
*{10}{c}
!{\color{black}\vrule width 0.4pt}
*{7}{c}
}
\toprule
\multirow{2}{*}{\textbf{Method}}
& \multicolumn{10}{
c!{\color{black}\vrule width 0.4pt}
}{\textbf{Semantic Fidelity}}
& \multicolumn{7}{c}{\textbf{Internal Consistency}} \\
\cmidrule(lr){2-11}
\cmidrule(lr){12-18}
& $\mathrm{GSF}\uparrow$ & $\mathrm{SAP}\uparrow$
& $\mathrm{ANF}\uparrow$ & $\mathrm{DLI}\uparrow$
& $\mathrm{ARP}\uparrow$ & $\mathrm{CCP}\uparrow$
& $\mathrm{LCF}\uparrow$ & $\mathrm{SSP}\uparrow$
& $\mathrm{OTI}\uparrow$
& $\boldsymbol{\mathit{Overall}}\uparrow$
& $\mathrm{SAC}\uparrow$ & $\mathrm{ANL}\uparrow$
& $\mathrm{TSC}\uparrow$ & $\mathrm{SPC}\uparrow$
& $\mathrm{PL}\uparrow$ & $\mathrm{AVS}\uparrow$
& $\boldsymbol{\mathit{Overall}}\uparrow$ \\
\midrule

\rowcolor{gray!18}
\multicolumn{18}{c}{\textit{T2V-PE}} \\
\rowcolor{white}
WanPE-397B & \underline{89.20} & 70.00 & 80.80 & \textbf{94.20} & 62.00 & \underline{57.40} & \underline{80.40} & 80.60 & 94.80 & \underline{74.89} & 91.20 & 87.40 & 94.60 & 82.40 & 92.80 & \underline{94.40} & 91.51 \\
WanPE-35B & 86.40 & 71.80 & \underline{82.60} & 88.00 & 59.60 & 54.40 & 73.20 & 81.20 & \textbf{97.00} & 72.83 & 91.60 & 89.20 & 92.40 & 81.60 & 91.60 & 90.20 & 91.40 \\
H3-Context-IR & 81.40 & 66.40 & 70.40 & \underline{93.00} & 58.40 & 49.20 & 76.80 & 81.20 & \underline{95.40} & 67.34 & \textbf{97.20} & \textbf{95.40} & \underline{96.80} & \textbf{98.00} & 91.80 & 90.80 & 94.50 \\
SCMaPR & \textbf{94.60} & \textbf{87.60} & \textbf{85.60} & 80.20 & \textbf{82.00} & \textbf{87.20} & \textbf{90.60} & \textbf{92.60} & 95.00 & \textbf{87.92} & \underline{94.60} & \underline{92.60} & \underline{96.80} & \underline{91.40} & \underline{93.60} & \textbf{95.60} & \underline{95.06} \\
LTX-2.5-PE & 83.00 & \underline{78.00} & 76.20 & 80.40 & \underline{75.20} & 43.00 & 72.20 & \underline{82.20} & 91.20 & 71.21 & 88.60 & 80.60 & 85.60 & 77.60 & 89.80 & 91.00 & 87.52 \\
LingBot-Video-PE & 67.80 & 28.00 & 30.40 & 72.00 & 54.00 & 16.80 & 28.00 & 45.40 & 84.60 & 41.61 & 91.20 & 91.00 & 93.80 & 66.60 & 92.40 & 93.20 & 90.96 \\
VPO & 74.00 & 65.00 & 45.60 & 76.20 & 63.40 & 45.00 & 66.00 & 66.40 & 89.00 & 62.80 & 92.60 & 90.00 & \textbf{97.20} & \underline{91.40} & 93.00 & 94.00 & \textbf{95.07} \\
Prompt-A-Video-Cog & 84.60 & 47.60 & 35.40 & 78.60 & 69.20 & 29.00 & 58.60 & 44.40 & 91.20 & 56.48 & 82.40 & 81.80 & 93.40 & 86.20 & 89.40 & 89.80 & 91.09 \\
Prompt-A-Video-OS & 81.20 & 46.60 & 30.40 & 76.80 & 67.20 & 33.20 & 56.20 & 48.00 & 89.80 & 54.83 & 86.00 & 85.80 & 96.40 & 90.00 & 93.40 & 93.00 & 93.87 \\
RAPO & 37.20 & 6.40 & 1.00 & 72.60 & 52.40 & 24.20 & 25.40 & 4.80 & 79.00 & 15.47 & 80.80 & 64.00 & 88.20 & 76.80 & \textbf{93.80} & 93.80 & 88.30 \\
\midrule

\rowcolor{gray!18}
\multicolumn{18}{c}{\textit{I2V-PE}} \\
\rowcolor{white}
H3-Context-IR & 64.67 & \textbf{56.67} & \underline{67.00} & \textbf{89.00} & 47.33 & \textbf{52.00} & \textbf{87.67} & \textbf{76.33} & \textbf{81.00} & \textbf{58.59} & \textbf{97.67} & \textbf{96.33} & \textbf{97.33} & \textbf{99.33} & \underline{89.00} & \underline{94.33} & \textbf{95.05} \\
LTX-2.5-PE & \textbf{84.67} & \underline{56.33} & \textbf{71.00} & \underline{81.67} & \textbf{65.00} & \underline{24.00} & \underline{78.00} & 72.00 & \underline{71.67} & \underline{56.77} & 85.67 & \underline{80.00} & 88.00 & \underline{76.33} & 86.33 & 90.33 & 87.36 \\
LingBot-Video-PE & \underline{72.67} & 54.33 & 23.33 & 77.00 & \underline{57.33} & 10.67 & 42.33 & \underline{73.00} & 70.67 & 45.70 & \underline{95.67} & \textbf{96.33} & \underline{90.00} & 66.67 & \textbf{89.67} & \textbf{95.00} & \underline{91.14} \\
\midrule

\rowcolor{gray!18}
\multicolumn{18}{c}{\textit{R2V-PE}} \\
\rowcolor{white}
H3-Context-IR & \textbf{73.33} & \textbf{54.00} & \textbf{65.00} & \textbf{83.33} & \underline{56.67} & \textbf{51.00} & \textbf{76.33} & \textbf{73.33} & \textbf{82.00} & \textbf{58.11} & \textbf{87.33} & \textbf{90.00} & \textbf{98.67} & \textbf{95.33} & \textbf{94.67} & \textbf{91.67} & \textbf{94.90} \\
LTX-2.5-PE & \underline{61.67} & \underline{43.67} & \underline{62.67} & \underline{70.00} & \textbf{71.00} & \underline{27.67} & \underline{64.00} & \underline{58.00} & \underline{75.67} & \underline{49.03} & \underline{76.33} & \underline{79.33} & \underline{88.67} & \underline{74.67} & \underline{93.67} & \underline{82.67} & \underline{86.66} \\
\bottomrule
\end{tabular}%
}

\vspace{5pt}

\setlength{\tabcolsep}{8pt}

\resizebox{\linewidth}{!}{%
\begin{tabular}{
l!{\color{black}\vrule width 0.4pt}
*{10}{c}
}
\toprule
\multirow{2}{*}{\textbf{Method}}
& \multicolumn{10}{c}{\textbf{Directorial Creation}} \\
\cmidrule(lr){2-11}
& $\mathrm{STC}\uparrow$ & $\mathrm{CDS}\uparrow$
& $\mathrm{ETD}\uparrow$ & $\mathrm{LCS}\uparrow$
& $\mathrm{NDE}\uparrow$ & $\mathrm{CIF}\uparrow$
& $\mathrm{SDI}\uparrow$ & $\mathrm{SPS}\uparrow$
& $\mathrm{VSS}\uparrow$
& $\boldsymbol{\mathit{Overall}}\uparrow$ \\
\midrule

\rowcolor{gray!18}
\multicolumn{11}{c}{\textit{T2V-PE}} \\
\rowcolor{white}
WanPE-397B & \textbf{66.60} & \underline{81.00} & \underline{63.20} & \textbf{80.20} & \underline{68.60} & 91.20 & 77.60 & 81.60 & \textbf{82.20} & \textbf{78.97} \\
WanPE-35B & \underline{59.80} & 80.20 & \textbf{63.60} & 75.60 & 66.60 & \underline{94.20} & \underline{83.00} & 80.40 & \underline{78.60} & \underline{78.15} \\
H3-Context-IR & 12.20 & \textbf{89.00} & 55.40 & \underline{76.80} & \textbf{76.40} & 94.00 & \textbf{93.60} & \textbf{86.40} & 73.60 & 76.73 \\
SCMaPR & 6.20 & 55.20 & 41.60 & 48.60 & 61.00 & \textbf{99.20} & 13.20 & 74.80 & 70.60 & 49.18 \\
LTX-2.5-PE & 2.40 & 76.60 & 36.40 & 66.80 & 65.40 & \underline{94.20} & 61.20 & \underline{82.60} & 77.40 & 57.69 \\
LingBot-Video-PE & 40.60 & 62.80 & 46.20 & 35.60 & 60.20 & 65.80 & 4.60 & 77.80 & 70.40 & 48.66 \\
VPO & 0.40 & 25.20 & 27.40 & 64.20 & 48.80 & 89.60 & 6.60 & 75.20 & 70.00 & 33.07 \\
Prompt-A-Video-Cog & 0.60 & 23.00 & 21.00 & 58.20 & 48.40 & 83.80 & 14.60 & 65.20 & 71.20 & 29.66 \\
Prompt-A-Video-OS & 1.00 & 19.00 & 14.40 & 30.80 & 22.40 & 86.80 & 9.60 & 37.40 & 54.40 & 18.99 \\
RAPO & 0.00 & 1.40 & 1.80 & 2.40 & 1.40 & 14.40 & 1.00 & 3.60 & 4.20 & 0.93 \\
\midrule

\rowcolor{gray!18}
\multicolumn{11}{c}{\textit{I2V-PE}} \\
\rowcolor{white}
H3-Context-IR & \textbf{34.33} & \textbf{73.91} & \textbf{75.92} & \textbf{79.60} & \textbf{87.96} & \underline{82.94} & \textbf{96.99} & \textbf{75.25} & 61.20 & \textbf{73.35} \\
LTX-2.5-PE & 2.00 & 43.67 & \underline{52.33} & \underline{73.33} & \underline{53.33} & \textbf{91.00} & \underline{60.33} & 65.33 & \underline{75.00} & \underline{46.77} \\
LingBot-Video-PE & \underline{19.00} & \underline{50.00} & 48.33 & 43.00 & 38.33 & 72.00 & 5.33 & \underline{67.67} & \textbf{83.00} & 37.47 \\
\midrule

\rowcolor{gray!18}
\multicolumn{11}{c}{\textit{R2V-PE}} \\
\rowcolor{white}
H3-Context-IR & \textbf{23.00} & \textbf{78.00} & \textbf{76.67} & \textbf{80.33} & \textbf{93.33} & \textbf{92.33} & \textbf{84.00} & \textbf{68.00} & \textbf{70.33} & \textbf{73.62} \\
LTX-2.5-PE & \underline{5.67} & \underline{49.67} & \underline{41.67} & \underline{62.00} & \underline{62.00} & \underline{81.00} & \underline{51.00} & \underline{42.33} & \underline{62.67} & \underline{38.02} \\
\bottomrule
\end{tabular}%
}

\label{tab:main_results}
\end{table*}

\subsection{Main Results}
Table~\ref{tab:main_results} reports PE performance across all three tasks.
On T2V-PE, different methods lead in three dimensions, while WanPE-397B presents a balanced profile by ranking first in Directorial Creation and second in Semantic Fidelity. On the visually conditioned tasks, H3-Context-IR ranks first in all three dimensions on both I2V-PE and R2V-PE.
No single method therefore dominates across all tasks and dimensions, revealing distinct capability profiles among current PE.

\textbf{Finding 1:} \textit{Modern prompt enhancement is evolving from fine-grained descriptive expansion toward intent-preserving cinematic planning, enabled by increasingly capable video generators.}

Modern video generators can follow richer, more structured instructions for temporal progression, camera design, narrative development, and audiovisual coordination within a coherent generation plan, expanding the role of PE beyond local detail enrichment. 
This shift is evident in large-scale, industry-developed PE: WanPE establishes a clear lead in Directorial Creation on T2V-PE, while H3-Context-IR leads by substantial margins on I2V-PE and R2V-PE.
Appendix~\ref{app:all_task_cases} contrasts descriptive expansion with structured cinematic planning across PE systems.
Beyond this emerging industrial trend, Internal Consistency remains uniformly high, varying by less than 9 points within each task, whereas Semantic Fidelity and Directorial Creation each span more than 70 points on T2V-PE.
These results indicate that the central challenge lies not merely in expanding prompts or maintaining internal coherence, but in cinematic planning that preserves user intent. 

\textbf{Finding 2:} \textit{PE trained to reconstruct real-video captions exhibits broader coverage of cinematic language, while forward-refinement methods excel at semantic preservation or internal coherence.}

The dimension-level results on T2V-PE reveal a clear correspondence between enhancement paradigms and capability profiles.
WanPE-397B and WanPE-35B are trained on pairs constructed by annotating real videos with a video captioner and reconstructing compatible user requests from the resulting captions, while retaining the full captions as targets. 
This supervision exposes them to real-world temporal organization, cinematic language, and audiovisual coordination, consistent with their leading Directorial Creation scores of 78.97 and 78.15 and their strong performance in storyboard and timing, editing and transitions, and sound design.
In contrast, SCMaPR combines policy-conditioned rewriting, atom-level semantic verification, and conditional revision, whereas VPO trains a prompt optimizer through principle-based SFT and multi-feedback preference optimization.
These mechanisms help preserve user requirements and constrain unsupported or contradictory elaboration, consistent with their respective leadership in Semantic Fidelity (87.92) and Internal Consistency (95.07), but their lower Directorial Creation scores (49.18 and 33.07) indicate narrower cinematic coverage.
These complementary profiles suggest that effective PE should combine caption-derived planning knowledge with explicit semantic and logical constraints.

\begin{table*}[t]
\centering
\begin{minipage}[t]{0.48\textwidth}
\centering
\caption{Temporal-complexity results.}
\label{tab:temporal_complexity}
\small
\renewcommand{\arraystretch}{1}
\setlength{\tabcolsep}{8pt}
\resizebox{\linewidth}{!}{%
\begin{tabular}{ccccc}
\toprule
Structure & Count & SF & IC & DC \\
\midrule
Event stages
  & 1  & 69.60 & 96.66 & 78.43 \\
  & 2  & 68.79 & 94.02 & 76.70 \\
  & 3+ & 65.93 & 92.74 & 76.45 \\
\midrule
Requested shots
  & 1  & 72.48 & 95.81 & 79.03 \\
  & 2  & 69.90 & 94.52 & 75.62 \\
  & 3+ & 62.51 & 88.20 & 73.51 \\
\bottomrule
\end{tabular}%
}
\end{minipage}
\hfill
\begin{minipage}[t]{0.48\textwidth}
\centering
\caption{Reference-complexity results.}
\label{tab:reference_complexity}
\small
\renewcommand{\arraystretch}{1}
\setlength{\tabcolsep}{8pt}
\resizebox{\linewidth}{!}{%
\begin{tabular}{ccccc}
\toprule
Reference & Group & SF & IC & DC \\
\midrule
Images
  & 1  & 62.01 & 96.70 & 77.09 \\
  & 2  & 60.37 & 95.57 & 76.56 \\
  & 3+ & 54.59 & 92.31 & 74.38 \\
\midrule
Modality
  & Image-only  & 58.64 & 94.66 & 75.88 \\
  & Video-only  & 67.79 & 94.46 & 74.75 \\
  & Hybrid & 45.62 & 97.09 & 55.69 \\
\bottomrule
\end{tabular}%
}
\end{minipage}

\vspace{-3mm}
\end{table*}

\subsection{Analysis}
Building on the preceding findings, we analyze the necessity of prompt enhancement, the relationship between output length and adequacy, and the impact of task complexity.

\textbf{Original prompts reveal a substantial adequacy gap.}
We evaluate the original prompts as a no-enhancement baseline. Across all tasks, they achieve high Semantic Fidelity but low Directorial Creation. On T2V-PE, they attain 100.00 in Semantic Fidelity but only 33.56 in Directorial Creation, indicating that preserving user intent alone is insufficient to fully specify the intended video.
This observation raises the question of whether greater prompt length necessarily yields higher Directorial Creation. A length-stratified analysis shows no consistent benefit: across tasks, the longest output quartile trails the shortest by 5.19-10.09 points (see Appendix~\ref{app:length_analysis} for details).
Effective PE should therefore add structured, task-relevant specifications for temporal planning, cinematography, narrative development, and audiovisual design rather than merely lengthening prompts.

\textbf{Finding 3:} \textit{Temporal and reference complexity remain persistent challenges for prompt enhancers.}

To examine these challenges, we use H3-Context-IR as a fixed enhancer and analyze two forms of task complexity: temporal complexity on T2V-PE and reference complexity on R2V-PE.

\textbf{Temporal-structure complexity.}
We analyze T2V-PE to isolate temporal complexity from visual-reference grounding. Specifically, we characterize temporal structure by the numbers of ordered event stages and explicitly requested shots, grouping both into 1, 2, and 3+. An event stage denotes a distinct action, state change, or outcome, while the shot analysis includes only prompts with explicit shot structures.
Under these groupings, Table~\ref{tab:temporal_complexity} shows that cases with three or more event stages trail those with one stage by 3.67, 3.92, and 1.98 points in SF, IC, and DC, respectively. The corresponding gaps for requested shots are larger at 9.97, 7.61, and 5.52 points.
Together, these results show that both event progression and multi-shot organization pose temporal challenges, with the latter placing greater demands on cross-segment coordination.

\textbf{Reference complexity.}
On R2V-PE, we characterize reference complexity along two axes: quantity and modality. For quantity, we group image-only cases by 1, 2, and 3+ images. For modality, we compare image-only, video-only, and hybrid inputs.
As shown in Table~\ref{tab:reference_complexity}, cases with three or more images trail single-image cases by 7.42, 4.39, and 2.71 points in SF, IC, and DC, respectively. The modality comparison reveals a larger gap. Hybrid inputs score only 45.62 in SF and 55.69 in DC, substantially below both single-modality settings, despite achieving the highest IC of 97.09.
This divergence reveals a reference-grounding bottleneck. Enhanced prompts can remain internally coherent while failing to faithfully and adequately integrate heterogeneous references.

\begin{figure*}[t]
\centering  
    \begin{overpic}[width=1\linewidth]{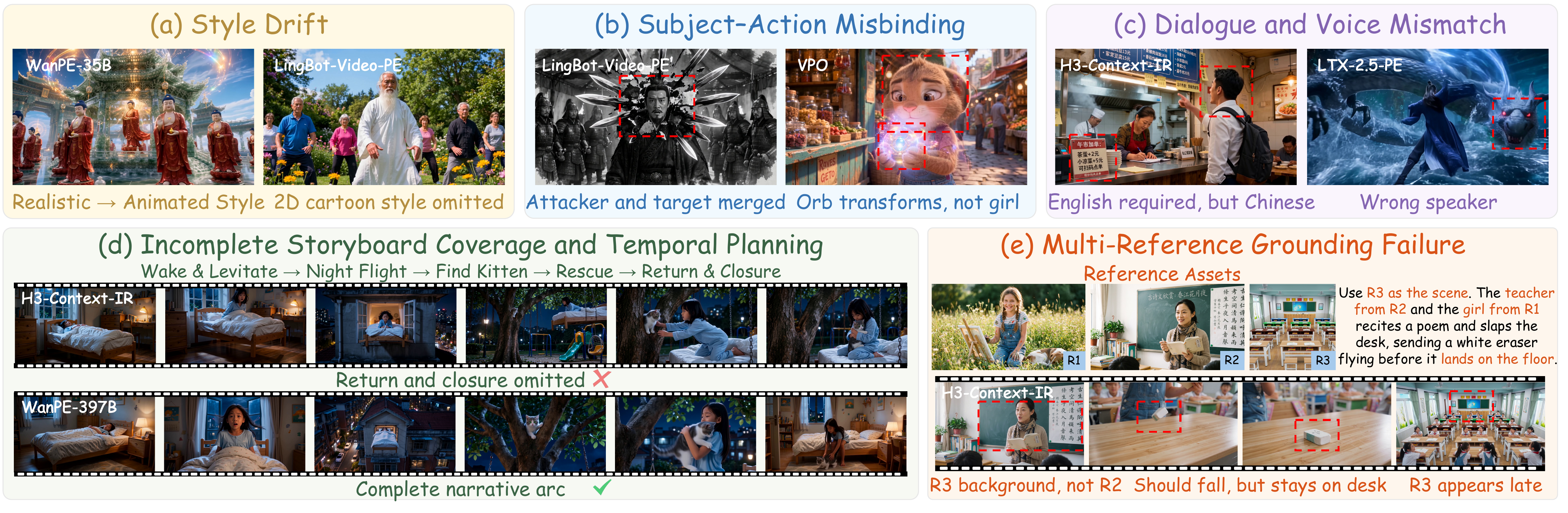}
    \end{overpic} 
    \caption{\textbf{Qualitative failure cases of evaluated prompt enhancers, visualized with Wan3.0 video model.} Examples include (a) visual-style deviation, (b) incorrect subject–action binding, (c) language and speaker attribution errors, (d) narrative truncation caused by inadequate temporal planning, and (e) incorrect grounding across multiple references. }
    \label{fig:fail_case}
\end{figure*}

\begin{table*}[t]
\centering
\caption{Spearman correlations for human alignment and downstream utility across three tasks.}
\label{tab:human_downstream_alignment}
\small
\renewcommand{\arraystretch}{1}
\setlength{\tabcolsep}{12pt}
\resizebox{\linewidth}{!}{%
\begin{tabular}{lccccccccc}
\toprule
& \multicolumn{3}{c}{T2V-PE}
& \multicolumn{3}{c}{I2V-PE}
& \multicolumn{3}{c}{R2V-PE} \\
\cmidrule(lr){2-4}
\cmidrule(lr){5-7}
\cmidrule(lr){8-10}
Evaluation & SF & IC & DC & SF & IC & DC & SF & IC & DC \\
\midrule
\rowcolor{gray!18}
\multicolumn{10}{c}{\textit{Enhanced-prompt alignment}} \\
\rowcolor{white}
Direct VLM Scoring
& 0.618 & 0.464 & 0.527 & 1.000 & 0.600 & 0.400 & 1.000 & 0.500 & 0.500 \\
\bench{} Evaluation
& 0.955 & 0.818 & 0.909 & 1.000 & 1.000 & 1.000 & 1.000 & 1.000 & 1.000 \\
\midrule
\rowcolor{gray!18}
\multicolumn{10}{c}{\textit{Downstream-video alignment}} \\
\rowcolor{white}
\bench{} Evaluation (Wan3.0)
& 0.845 & 0.827 & 0.945 & 1.000 & 1.000 & 1.000 & 1.000 & 1.000 & 1.000 \\
\bench{} Evaluation (MiniMax-H3)
& 0.809 & 0.864 & 0.936 & 1.000 & 1.000 & 1.000 & 1.000 & 1.000 & 1.000 \\
\bottomrule
\end{tabular}
}
\end{table*}

\textbf{Qualitative failure analysis.}
As shown in Figure~\ref{fig:fail_case}, evaluated PE exhibit representative failures across all three dimensions. The corresponding videos generated by Wan3.0 visibly reflect these errors, illustrating the interpretability and downstream relevance of our prompt-level evaluation. 

\subsection{Human Alignment}
\textbf{Finding 4:} \textit{\bench{} evaluation scores reliably reflect expert judgments of both enhanced prompts and downstream videos across generators.}

To validate the reliability of our evaluation framework, we measure Spearman correlations ($\rho_s$) with expert rankings at the enhanced-prompt and downstream-video levels. The resulting correlations are reported in Table~\ref{tab:human_downstream_alignment}.
At the enhanced-prompt level, our framework aligns more closely with expert rankings than direct VLM scoring, with dimension-averaged gains of 0.333-0.358 across tasks. This improvement supports the effectiveness of rubric-based decomposition.
At the downstream-video level, we compare the same evaluation scores with expert rankings of downstream videos generated by Wan3.0 and MiniMax-H3. The resulting correlations of 0.809-1.000 indicate that direct prompt evaluation reliably reflects downstream utility across generators.
The study covers all PE systems in Table~\ref{tab:main_results}, with the original prompts as a baseline. Annotation details are provided in Appendix~\ref{app:human_evaluation}.

%% file: section/conclusion.tex
\section{Conclusion}
We present \bench{}, the first unified benchmark for direct evaluation of prompt enhancers across T2V-PE, I2V-PE, and R2V-PE.
It comprises 1,100 expert-verified cases and 1,005 visual assets spanning 35 fine-grained tasks encompassing diverse temporal, cinematic, audiovisual, and multi-reference requirements.
Its evidence-grounded framework combines modality-aware evidence with 24 criterion-specific rubrics, enabling consistent, generator-independent assessment without costly video rendering.
Our evaluation reveals an emerging shift from descriptive expansion toward intent-preserving cinematic planning.
At the method level, caption-reconstruction and forward-refinement methods exhibit complementary strengths, suggesting that future PE systems should combine cinematic knowledge with semantic and logical safeguards.
Meanwhile, performance degradation under temporal and multi-reference complexity highlights persistent challenges in temporal planning and cross-asset grounding.
Strong alignment with expert judgments at both the prompt and downstream-video levels confirms the practical relevance of direct PE evaluation.
By unifying task coverage, evidence-grounded assessment, and criterion-level analysis, \bench{} provides a scalable foundation for evaluating, comparing, and advancing video prompt enhancers.

%% file: section/appendix.tex
\clearpage
\appendix

\newtcblisting{promptbox}[1]{
  enhanced,
  breakable,
  listing only,
  title={#1},
  colback=white,
  colframe=black!75,
  colbacktitle=black!75,
  coltitle=white,
  fonttitle=\bfseries,
  boxrule=0.6pt,
  arc=1mm,
  left=1.5mm,
  right=1.5mm,
  top=1mm,
  bottom=1mm,
  before skip=6pt,
  after skip=8pt,
  listing options={
    basicstyle=\ttfamily\scriptsize,
    breaklines=true,
    breakatwhitespace=false,
    columns=fullflexible,
    keepspaces=true,
    showstringspaces=false,
    escapeinside={(*@}{@*)}
  }
}

\section*{Appendix Overview}

This appendix supplements the main paper with data, evaluation, experimental,
and human annotation details, followed by ethical considerations and
limitations. It is organized as follows.

\begin{itemize}[leftmargin=*,itemsep=1pt,topsep=2pt]
    \item \textbf{Appendix A: More Data Details.} Data design, construction,
    and detailed analysis.
    \item \textbf{Appendix B: More Evaluation Details.} Evaluation dimensions
    and criteria, structured output formats, and representative case.
    \item \textbf{Appendix C: Additional Experimental Details.} Implementation
    details and length-stratified analysis.
    \item \textbf{Appendix D: Human Expert Annotation.} Pairwise annotation
    protocol, quality control, and annotation interfaces.
    \item \textbf{Appendix E: Ethics, Privacy, and Licensing.} Data screening,
    asset licensing, and release policy.
    \item \textbf{Appendix F: Limitations.} Benchmark scale and future
    expansion.
\end{itemize}

\section{More Data Details}
\label{app:more_data_details}

\subsection{Data Design Details}
\label{app:data_design}

\noindent\textbf{Prompt enhancement settings.}
The benchmark unifies three settings with progressively richer conditioning.
T2V-PE contains 500 text-only cases. I2V-PE contains 300 cases, each paired
with one first-frame image. R2V-PE contains 300 cases conditioned on one or
more images or videos whose roles may include subject identity, object,
scene, style, motion, cinematography, transition pattern, or audiovisual
rhythm. Together, these settings cover prompt enhancement from text, 
temporal continuation from a first frame, and integration of multiple references.

\noindent\textbf{Tasks.}
The benchmark defines a two-level task taxonomy with 9 high-level
capability categories and 35 fine-grained task types. Each task type belongs
to one parent category. Table~\ref{tab:task_taxonomy} lists the
taxonomy and the number of task types in each category.

\begin{table}[h!]
\centering
\small
\caption{\textbf{Hierarchical task taxonomy of \bench{}.} The nine
high-level categories contain 35 fine-grained task types.}
\label{tab:task_taxonomy}
\setlength{\tabcolsep}{4pt}
\renewcommand{\arraystretch}{1.10}
\resizebox{\linewidth}{!}{%
\begin{tabular}{lcp{0.73\linewidth}}
\toprule
High-level category & \# & Fine-grained task types \\
\midrule
Action & 4 & Daily Activity, Combat, Sports, Dance \\
Animation & 2 & 2D Animation, 3D Animation \\
Visual Effects & 5 & VFX Character, VFX Object, Transformation,
Light/Weather Effects, Sci-Fi Creature \\
General & 5 & Physical Laws, On-Screen Text, Portrait, Nature,
Intellectual Property (IP) \\
Narrative & 3 & Modern Story, Historical Story, Fantasy Story \\
Cinematography & 3 & Camera Motion, Time-Lapse, Lighting and Color \\
Speech & 5 & Dialogue, Monologue, Narration, Explanation,
Emotional Vocalization \\
Music & 6 & Singing, Song and Dance, Music Video, Opera,
Background Music, Instrumental Performance \\
Sound Effects & 2 & Animal Sound, Environmental Sound \\
\bottomrule
\end{tabular}%
}
\end{table}

\noindent\textbf{Scenario composition dimensions.}
Each scenario is composed along four additional axes. The five subject
categories are Human, Animal, Object, Fantasy Being, and No Salient Subject.
The eight scene domains are Domestic, Urban, Workplace, Performance, Sports,
Transportation, Nature, and Otherworldly. The seven visual style classes are
Photorealistic, Anime, Graphic Illustration, 3D CGI, Painting, Digital
Stylized, and Mixed/Unspecified. The eight explicit audio domains
are Speech and Vocalization, Singing, Music, Human Activity Sounds, Foley and
Object Sounds, Natural and Animal Sounds, Environmental Ambience, and
Mechanical and Vehicle Sounds.

\noindent\textbf{Linguistic, temporal, and cinematic dimensions.}
The multilingual portion spans English, Chinese, Japanese, Korean, Spanish,
French, Arabic, Russian, Bengali, German, Portuguese, and Hindi. Prompts vary
from concise concepts to detailed scripts and may describe a single
continuous shot or as many as 13 explicitly structured shots. Cinematic
coverage includes near, medium, and long shot scale families. It also covers
level, high, low, overhead or aerial, and subjective or special viewpoints.
Lens families include wide, standard, telephoto, and specialty lenses.
Camera motion includes translational, rotational, and compound movement.
Composition, focus, transition, lighting, palette, and color grading provide
additional controls that may overlap.

\subsection{Data Construction Details}
\label{app:data_construction_details}

Figure~\ref{fig:data_construct} summarizes the four-stage construction
pipeline. This section describes the design choices and quality checks used
at each stage.

\textbf{Stage 1: Taxonomy Design and Composition}

Domain experts define the nine capability categories and 35 tasks in
Table~\ref{tab:task_taxonomy}. The taxonomy covers visual, narrative,
cinematic, speech, music, and sound capabilities. It therefore tests both
frame content and the development of events, camera behavior, and audio over
time. The experts also define the subject, scene, style, and audio
vocabularies described in Section~\ref{app:data_design}. Each candidate
scenario is represented as a
$(\text{category},\text{subject},\text{scene},\text{style},\text{audio})$
quintuple. It is then paired with a task, a prompt enhancement setting, and a
target duration. Incompatible combinations are discarded rather than forced.
Coverage is monitored across the dataset to maintain diversity without
producing incoherent or infeasible requests.

\textbf{Stage 2: Prompt Synthesis and Diversification}

Gemini 3.1 Pro synthesizes a candidate user request from the sampled
specification. It considers the task and conditioning modality, the scenario
composition, and the content that can reasonably unfold within the target
duration. Instruction specificity ranges from short concepts to descriptive
requests and detailed scripts. Temporal structure also varies. Some requests
leave the shot structure implicit, while others specify ordered events,
timestamps, or multiple shots. This process varies how much information and
temporal planning are provided to the prompt enhancer.

\textbf{Stage 3: Reference Asset Acquisition}

Cases with reference inputs use two acquisition routes. Public data selection
draws candidate images and videos from existing datasets. Images are screened for visual
quality, content safety, resolution, and redundancy. Videos are screened for
visual quality and motion prominence, segmented into semantically complete
clips, and represented by keyframes when a static reference is needed.

Generation with models expands the public pool using Wan-Image~\citep{wanimage}, Seedream~\citep{seedream},
and GPT Image 2~\citep{gptimage2}. In the image guided route, a primary image is used to create
variations in viewpoint, pose, or expression while preserving identity. In
the text guided route, Gemini 3.1 Pro converts the sampled scenario or a
relevant shot description into a concise image generation instruction.
Together, the two routes support single image, multiple image, video only,
and mixed image and video conditioning. Each retained reference is assigned
an explicit role such as subject, scene, style, motion, or cinematography.

\textbf{Stage 4: Automated Verification and Expert Curation}

Quality control combines automated screening with expert review. Gemini
3.1 Pro derives a checklist from each request and its target duration. It
checks task alignment, logical coherence, and duration feasibility. For cases
with reference inputs, it jointly inspects the instruction and
candidate assets to verify their assigned roles, alignment with the script,
reference coverage, and consistency across assets.

Ten domain experts review the verified pool. They refine awkward
language, remove ambiguous or infeasible instructions, reject uninformative
or mismatched references, and eliminate repetitive or redundant cases. They
also monitor task balance, difficulty, and novelty across cases. The final
benchmark contains 1,100 cases verified by experts and 1,005 visual assets.

\subsection{Detailed Data Analysis}
\label{app:detailed_data_analysis}

\noindent\textbf{Task coverage.}
\bench{} covers 9 high level capability categories: Action (25.3\%), Visual
Effects (16.0\%), Narrative (13.4\%), Animation (10.0\%), Speech (9.7\%),
General (9.1\%), Music (7.6\%), Cinematography (6.8\%), and Sound Effects
(2.1\%).

\noindent\textbf{Subject, scene, and style diversity.}
Subjects span 5 categories: Human (50.4\%), Object (16.6\%), Fantasy Being
(14.4\%), Animal (10.1\%), and No Salient Subject (8.5\%). Scenes span 8
categories: Otherworldly (19.8\%), Domestic (14.5\%), Urban (14.1\%),
Workplace (14.1\%), Nature (13.2\%), Performance (11.9\%), Sports (7.5\%),
and Transportation (5.0\%). T2V-PE visual styles span 7 categories:
Photorealistic (61.2\%), Digital Stylized (10.4\%), 3D CGI (8.2\%),
Mixed/Unspecified (8.2\%), Graphic Illustration (5.0\%), Painting (3.8\%),
and Anime (3.2\%).

\noindent\textbf{Audio and language diversity.}
Prompts cover 12 languages: English (50.0\%), Chinese (40.0\%), Japanese
(1.0\%), Korean (1.0\%), Spanish (1.0\%), French (1.0\%), Arabic (1.0\%),
Russian (1.0\%), Bengali (1.0\%), German (1.0\%), Portuguese (1.0\%), and
Hindi (1.0\%). Audio content spans 8 categories: Speech and
Vocalization (38.3\%), Music (19.9\%), Foley
and Object Sounds (14.6\%), Singing (12.6\%), Natural and Animal Sounds
(7.7\%), Mechanical and Vehicle Sounds (3.1\%), Human Activity Sounds
(2.7\%), and Environmental Ambience (1.1\%).

\noindent\textbf{Prompt length.}
Prompt length spans 7 intervals: at most 50 tokens (14.1\%), 51 to 100 tokens
(13.8\%), 101 to 200 tokens (17.7\%), 201 to 400 tokens (20.1\%), 401 to 800
tokens (16.4\%), 801 to 1,600 tokens (15.9\%), and more than 1,600 tokens
(2.0\%).
Table~\ref{tab:prompt_length_statistics} reports detailed token length
statistics for each task.
Target durations range from 2 to 30 seconds. There are 147 cases from 2 to 5
seconds, 321 from 6 to 10 seconds, 614 from 11 to 15 seconds, 11 from 16 to 20
seconds, and 7 from 21 to 30 seconds.

\begin{table}[h!]
\centering
\small
\caption{\textbf{Input prompt token length statistics.} Quartiles are
computed independently within each setting.}
\label{tab:prompt_length_statistics}
\setlength{\tabcolsep}{15pt}
\renewcommand{\arraystretch}{1.08}
\resizebox{\linewidth}{!}{%
\begin{tabular}{lrrrrrr}
\toprule
Setting & Min. & Q1 & Median & Mean & Q3 & Max. \\
\midrule
T2V-PE & 6 & 85.3 & 237.5 & 425.8 & 597.3 & 3,431 \\
I2V-PE & 8 & 91.8 & 221.5 & 448.9 & 642.8 & 4,120 \\
R2V-PE & 13 & 89.8 & 210.5 & 363.8 & 549.0 & 2,412 \\
\midrule
All & 6 & 88.8 & 224.0 & 415.2 & 596.8 & 4,120 \\
\bottomrule
\end{tabular}
}
\end{table}

\noindent\textbf{Cinematic language.}
\bench{} reports 3 major shot scale families: near (49.5\%), long (30.7\%),
and medium (19.8\%). It covers 5 major camera angles: eye level (38.0\%),
high (23.2\%), low (20.2\%), subjective (13.7\%), and others (4.9\%). It also
covers 4 lens categories: wide angle (38.1\%), telephoto or zoom (27.6\%),
specialty (25.4\%), and standard (9.0\%). Camera movement spans 3 types: pure
translation (50.7\%), compound (44.2\%), and pure rotation (5.0\%).

\noindent\textbf{Reference diversity and complexity.}
R2V-PE reference configurations comprise 3 types: image only (80.0\%), video
only (9.3\%), and mixed image and video (10.7\%). Cases with multiple images
span 6 reference counts: two (39.2\%), three (21.7\%), four (19.3\%), five
(16.9\%), six (1.8\%), and seven (1.2\%). Multi-image reference structures
comprise 5 types: multi view (5.4\%), multiple subjects (15.1\%), subject with
object or scene (56.0\%), sequential frames (21.7\%), and others (1.8\%).
Reference asset styles comprise 3 groups: realistic (54.6\%), nonrealistic
(24.8\%), and mixed or unspecified (20.6\%).
In absolute terms, I2V-PE contributes 300 first frame images. R2V-PE
contributes 644 images and 61 videos. Its image conditioned cases include 106
with one image, 65 with two, 36 with three, 32 with four, 28 with five, 3 with
six, and 2 with seven. Its video conditioned cases include 59 with one video
and 1 with two videos.

\section{More Evaluation Details}
\label{app:evaluation_details}

\subsection{Evaluation Dimensions and Criteria}
\label{app:evaluation_dimensions}

\bench{} evaluates enhanced prompts using 24 criteria grouped into Semantic
Fidelity (\textit{SF}), Internal Consistency (\textit{IC}), and Directorial
Creation (\textit{DC}). For each applicable criterion, the evaluator determines
whether the enhanced prompt contains no issue, a minor issue, or a major
issue, and cites evidence for that decision. A criterion is counted as passed
only when no issue is identified. Criterion-level pass rate is therefore the
percentage of applicable cases that pass the criterion, while each
dimension-level score aggregates the number and severity of its constituent
issues. Criteria tied to optional source content, such as dialogue or
on-screen text, are applied only when that content is present or explicitly
requested.

\subsubsection{Semantic Fidelity Metrics}
\label{app:sf_metrics}

Semantic Fidelity evaluates whether the enhanced prompt preserves the user
instruction and instruction-relevant visual evidence. Added details must not
omit, weaken, substitute, or contradict source requirements.

\noindent\textbf{(1) Genre and Style Fidelity (GSF).}
Measures preservation of the requested genre, medium, aesthetic, and visual
style. Replacing the medium, weakening a defining genre convention, or adding
an incompatible reference style constitutes an issue.

\noindent\textbf{(2) Subject and Attribute Preservation (SAP).}
Measures preservation of subject identity, count, appearance, attributes,
objects, and their bindings. I2V uses the first frame as the initial-state
anchor. R2V follows each reference's assigned role.

\noindent\textbf{(3) Action and Narrative Fidelity (ANF).}
Measures preservation of required actions, interactions, event order,
outcomes, and narrative intent. Omissions, substitutions, reversals, or added
events that alter the requested event chain are treated as issues.

\noindent\textbf{(4) Dialogue and Lyrics Integrity (DLI).}
Measures fidelity to dialogue, narration, lyrics, language, delivery, and
speaker attribution. Exact quotations must not be paraphrased, deleted,
translated, or converted into another form unless explicitly permitted.

\noindent\textbf{(5) Audio Requirement Preservation (ARP).}
Measures preservation of speech, vocalization, music, sound effects,
ambience, and silence constraints. Missing, substituted, or added sounds
that violate negative requirements such as ``no music'' are penalized.

\noindent\textbf{(6) Camera Constraint Preservation (CCP).}
Measures preservation of shot count, scale, angle, lens, focus, movement, and
single-take or static-camera constraints. Added camera design is allowed only
where it does not override explicit cinematic intent.

\noindent\textbf{(7) Lighting and Color Fidelity (LCF).}
Measures preservation of time of day, illumination, light quality,
brightness, palette, saturation, contrast, and grading. Compatible elaboration
is allowed, but it must not change a defining lighting or color condition.

\noindent\textbf{(8) Scene and Spatial Preservation (SSP).}
Measures preservation of setting, background elements, layout, relative
positions, and movement directions. Changing the location, removing required
objects, or reversing spatial relations is a violation.

\noindent\textbf{(9) On-Screen Text Integrity (OTI).}
Measures fidelity to visible titles, signs, subtitles, and interface text,
including exact wording, spelling, language, placement, and timing. Visible
text cannot be replaced by semantically similar spoken content.

\subsubsection{Internal Consistency Metrics}
\label{app:ic_metrics}

Internal Consistency evaluates whether the complete enhanced prompt is
logically, spatially, temporally, physically, and audiovisually coherent.

\noindent\textbf{(10) Subject and Attribute Continuity (SAC).}
Measures stability of subject identity, count, appearance, clothing, props,
and persistent attributes across shots. Unexplained renaming, duplication,
disappearance, or attribute drift constitutes an issue.

\noindent\textbf{(11) Action and Narrative Logic (ANL).}
Measures whether actions form a coherent causal or procedural progression.
The evaluator checks prerequisites, bindings between subjects and actions, reactions,
resolution, and incompatible or dangling events.

\noindent\textbf{(12) Temporal and State Continuity (TSC).}
Measures chronology, duration allocation, event order, and state transitions.
Overlapping or infeasible timestamps, unexplained resets, and later shots
showing states that should no longer exist are penalized.

\noindent\textbf{(13) Spatial Consistency (SPC).}
Measures consistency of location, screen direction, relative position,
orientation, trajectory, and entrances or exits. Unexplained position swaps
or movement paths that cannot connect adjacent states constitute issues.

\noindent\textbf{(14) Physical Logic (PL).}
Measures compliance with real-world physics or established fictional rules,
including gravity, contact, object permanence, bodily feasibility, and
causal relations. Deliberate fantasy is judged by its own world logic.

\noindent\textbf{(15) Audiovisual Synchronization (AVS).}
Measures temporal and semantic alignment between visible events and dialogue,
lyrics, music, effects, or ambience. Wrong speakers, premature effects, and
misaligned lip-sync or beat-sync actions are penalized.

\subsubsection{Directorial Creation Metrics}
\label{app:dc_metrics}

Even casually captured videos involve temporal progression, framing, camera
behavior, lighting, performance, visual style, and sound. A prompt may
preserve the source and remain logically coherent while leaving these
directorial elements underspecified. To assess whether a prompt enhancer
develops them into an actionable production plan, we introduce Directorial
Creation, covering storyboarding, cinematography, editing, lighting,
narrative, sound, performance, and visual style.

\noindent\textbf{(16) Storyboard and Timing Completeness (STC).}
Measures whether the content is organized into a feasible temporal
plan with adequate event or shot boundaries, ordering, and durations.

\noindent\textbf{(17) Camera Design Specificity (CDS).}
Measures actionable specs of framing, scale, angle, viewpoint, lens,
focus, and movement. 

\noindent\textbf{(18) Editing and Transition Design (ETD).}
Measures whether cuts and transitions support temporal, spatial, narrative,
or rhythmic progression. Disconnected shots or repeated transitions without
clear continuity or motivation are considered inadequate.

\noindent\textbf{(19) Lighting and Color Specification (LCS).}
Measures actionable specification of light source, direction, quality,
contrast, palette, saturation, and grading. 

\noindent\textbf{(20) Narrative Development and Engagement (NDE).}
Measures sufficient progression, variation, escalation, or resolution for
the requested content form. Static restatement and unmotivated repetition are
inadequate.

\noindent\textbf{(21) Core-Intent Focus (CIF).}
Measures whether additions support the central request. Repetitive padding,
template language, unnecessary subplots, and decorative details that distract
from the intended result reduce adequacy.

\noindent\textbf{(22) Sound Design Integration (SDI).}
Measures whether speech, music, effects, and ambience are integrated through
source, character, intensity, progression, and visual timing. It rewards
content-appropriate design rather than requiring every audio type.

\noindent\textbf{(23) Subject and Performance Specificity (SPS).}
Measures generation-relevant subject and performance detail, including
appearance, pose, expression, gesture, movement, interaction, gaze, and vocal
delivery. 

\noindent\textbf{(24) Visual Style Specification (VSS).}
Measures actionable specification of medium, texture, rendering language,
production design, and visual treatment.

\subsection{Structured Output Formats}
\label{app:evaluation_templates}

\subsubsection{Asset-Fact Extraction Output}

We present R2V as the representative example because it contains the most
complete multimodal evidence structure. Qwen3.5-Omni-Plus~\citep{qwenteam2026qwen35omnitechnicalreport} returns the following
fixed \textit{AssetFacts} structure, with each fact traceable to its source
asset.

\begin{promptbox}{R2V AssetFacts Output Format}
{
  "asset_inventory": {
    "total_count": 0,
    "image_count": 0,
    "video_count": 0,
    "items": [
      {
        "asset_id": "image_1|video_1",
        "asset_type": "image|video",
        "input_index": 1,
        "modality_index": 1
      }
    ]
  },
  "assets": [
    {
      "asset_id": "image_1|video_1",
      "asset_type": "image|video",
      "duration_seconds": "xxx",
      "media_and_style": {
        "media_type": "live_action|2d_animation|3d_cg|stop_motion|illustration|game_render|mixed",
        "realism": "realistic|stylized_realistic|non_realistic",
        "primary_style": "xxx [dominant visual style]",
        "secondary_styles": ["xxx [additional visual style]"],
        "visible_style_evidence": ["xxx [visible evidence for the style]"],
        "certainty": "certain|uncertain",
        "uncertainty_reason": "xxx [reason when uncertain]"
      },
      "overall_content_summary": "xxx [brief summary of the asset]",
      "subjects": [
        {
          "id": "image_1_subject_1|video_1_subject_1",
          "role": "primary|secondary|background_group",
          "category": "xxx [person, animal, object, or group]",
          "identity": "xxx [visible identity description]",
          "count_exact": 1,
          "count_estimate": "xxx",
          "appearance": ["xxx [visible appearance detail]"],
          "pose_and_current_state": ["xxx [visible pose or current state]"],
          "expression": "xxx [visible expression if identifiable]",
          "held_or_worn_items": ["xxx [held or worn item]"],
          "frame_position": "xxx [location within the frame]",
          "occlusion": "none|partial|heavy",
          "certainty": "certain|uncertain",
          "uncertainty_reason": "xxx [reason when uncertain]"
        }
      ],
      "scene_and_spatial_relations": {
        "scene_type": "xxx [visible scene description]",
        "indoor_outdoor": "indoor|outdoor|mixed",
        "foreground_elements": ["xxx [foreground element]"],
        "midground_elements": ["xxx [midground element]"],
        "background_elements": ["xxx [background element]"],
        "spatial_relations": [
          {
            "subject": "xxx [subject ID or entity]",
            "relation": "xxx [visible spatial relation]",
            "object": "xxx [related object or entity]",
            "certainty": "certain|uncertain"
          }
        ],
        "cropped_or_offscreen_information": ["xxx [cropped or offscreen cue]"]
      },
      "camera_and_composition": {
        "shot_scale": "extreme_close_up|close_up|medium_close_up|medium_shot|medium_full_shot|full_shot|long_shot|extreme_long_shot|varied",
        "camera_angle": "xxx [camera angle]",
        "camera_height": "xxx [camera height]",
        "view_characteristics": "wide_angle_feel|standard_view|telephoto_compression|fisheye|varied",
        "composition": ["xxx [composition characteristic]"],
        "visual_focus": "xxx [primary visual focus]",
        "depth_of_field": "shallow|moderate|deep|varied",
        "camera_motion": ["xxx [observed camera motion]"],
        "certainty": "certain|uncertain",
        "uncertainty_reason": "xxx [reason when uncertain]"
      },
      "lighting_and_color": {
        "light_sources": ["xxx [visible light source]"],
        "light_direction_and_effect": ["xxx [light direction and visible effect]"],
        "time_of_day_cue": "xxx [visible time-of-day cue if present]",
        "color_temperature": "warm|neutral|cool|mixed",
        "dominant_colors": ["xxx [dominant color]"],
        "accent_colors": ["xxx [accent color]"],
        "certainty": "certain|uncertain",
        "uncertainty_reason": "xxx [reason when uncertain]"
      },
      "onscreen_text": {
        "present": false,
        "items": [
          {
            "text": "xxx [visible text transcription]",
            "type": "subtitle|title|logo|sign|interface|other",
            "language": "xxx [text language]",
            "frame_position": "xxx [text location in the frame]",
            "time_range": {
              "start_seconds": "xxx",
              "end_seconds": "xxx"
            },
            "certainty": "certain|uncertain",
            "uncertain_fragments": ["xxx [uncertain text fragment]"]
          }
        ]
      },
      "segments": [
        {
          "segment_id": "video_1_segment_1",
          "shot_index": 1,
          "is_new_shot": true,
          "start_seconds": "xxx",
          "end_seconds": "xxx",
          "transition_in": "xxx [incoming transition if present]",
          "scene": "xxx [scene during the segment]",
          "subjects_present": ["xxx [visible subject ID]"],
          "visible_actions_and_state_changes": ["xxx [visible action or state change]"],
          "subject_interactions": ["xxx [visible subject interaction]"],
          "object_events": ["xxx [visible object event]"],
          "camera": {
            "shot_scale": "xxx [segment shot scale]",
            "angle": "xxx [segment camera angle]",
            "camera_height": "xxx [segment camera height]",
            "view_characteristics": "xxx [segment lens characteristic]",
            "composition": ["xxx [segment composition]"],
            "focus": "xxx [segment visual focus]",
            "depth_of_field": "xxx [segment depth of field]",
            "movement": {
              "type": "static|push_in|pull_out|pan|tilt|track|crane|orbit|zoom|handheld|compound",
              "direction": "xxx [movement direction]",
              "speed": "xxx [movement speed]",
              "amplitude": "xxx [movement amplitude]",
              "stability": "xxx [movement stability]",
              "tracking_subject": "xxx [tracked subject ID if applicable]",
              "start_frame": "xxx [movement start state]",
              "end_frame": "xxx [movement end state]"
            }
          },
          "shot_change_from_previous": {
            "shot_scale_change": "xxx [change in shot scale]",
            "angle_change": "xxx [change in camera angle]",
            "camera_position_change": "xxx [change in camera position]",
            "composition_change": "xxx [change in composition]",
            "focus_change": "xxx [change in visual focus]",
            "movement_change": "xxx [change in camera movement]",
            "scene_change": "xxx [change in scene]",
            "subject_spatial_change": "xxx [change in subject position]",
            "continuity_basis": ["xxx [evidence supporting continuity]"]
          },
          "visual_effects": ["xxx [visible effect]"],
          "onscreen_text_refs": ["xxx [related text item ID]"],
          "audio_event_refs": ["xxx [related audio event ID]"],
          "certainty": "certain|uncertain",
          "uncertainty_reason": "xxx [reason when uncertain]"
        }
      ],
      "subject_tracks": [
        {
          "subject_id": "video_1_subject_1",
          "first_segment": "video_1_segment_1",
          "last_segment": "video_1_segment_1",
          "stable_identity_anchors": ["xxx [stable identity cue]"],
          "actions_in_order": ["xxx [ordered subject action]"],
          "state_changes_in_order": ["xxx [ordered subject state change]"],
          "certainty": "certain|uncertain",
          "uncertainty_reason": "xxx [reason when uncertain]"
        }
      ],
      "audio": {
        "present": false,
        "speech": [
          {
            "id": "video_1_audio_1",
            "time_range": {
              "start_seconds": "xxx",
              "end_seconds": "xxx"
            },
            "speaker_subject_id": "xxx [speaker subject ID if identifiable]",
            "source_type": "visible_speaker|offscreen_speaker|voiceover|unknown",
            "language": "xxx [spoken language]",
            "transcript": "xxx [verbatim speech transcript]",
            "vocal_characteristics": ["xxx [audible vocal characteristic]"],
            "certainty": "certain|uncertain",
            "uncertainty_reason": "xxx [reason when uncertain]"
          }
        ],
        "music": [
          {
            "id": "video_1_music_1",
            "time_range": {
              "start_seconds": "xxx",
              "end_seconds": "xxx"
            },
            "description": ["xxx [audible music characteristic]"],
            "certainty": "certain|uncertain",
            "uncertainty_reason": "xxx [reason when uncertain]"
          }
        ],
        "ambient_sounds": ["xxx [audible ambient sound]"],
        "sound_effects": ["xxx [audible sound effect]"],
        "audio_visual_relations": ["xxx [audio-visual synchronization relation]"]
      },
      "uncertainties": ["xxx [asset-level uncertainty]"]
    }
  ],
  "cross_asset_relations": [
    {
      "source": {
        "asset_id": "image_1",
        "entity_id": "image_1_subject_1"
      },
      "target": {
        "asset_id": "video_1",
        "entity_id": "video_1_subject_1"
      },
      "relation": "same_entity|possibly_same_entity|same_scene|possibly_same_scene|same_object|possibly_same_object|style_similarity|other",
      "evidence": ["xxx [evidence for the cross-asset relation]"],
      "certainty": "certain|uncertain",
      "uncertainty_reason": "xxx [reason when uncertain]"
    }
  ],
  "asset_anchor_summary": ["xxx [instruction-relevant asset anchor]"],
  "uncertainties": ["xxx [global uncertainty]"]
}
\end{promptbox}

\subsubsection{Three-Dimension Evaluation Outputs}

The evaluator uses separate fixed outputs for Semantic Fidelity, Internal
Consistency, and Directorial Creation.

\begin{promptbox}{Semantic Fidelity Output Format}
{
  "overall_verdict": "good|acceptable|bad",
  "semantic_fidelity_score": "xxx [score from 0 to 100]",
  "primary_issue": "genre_style|subject|action_narrative|dialogue|sound|camera|lighting_color|scene|onscreen_text|none",
  "summary": "",
  "dimensions": {
    "genre_style":      {"status": "consistent|lossy|altered|not_present", "severity": "none|minor|major", "evidence_before": "", "evidence_after": "", "note": ""},
    "subject":          {"status": "consistent|lossy|altered|not_present", "severity": "none|minor|major", "evidence_before": "", "evidence_after": "", "note": ""},
    "action_narrative": {"status": "consistent|lossy|altered|not_present", "severity": "none|minor|major", "evidence_before": "", "evidence_after": "", "note": ""},
    "dialogue":         {"status": "consistent|lossy|altered|not_present", "severity": "none|minor|major", "evidence_before": "", "evidence_after": "", "note": ""},
    "sound":            {"status": "consistent|lossy|altered|not_present", "severity": "none|minor|major", "evidence_before": "", "evidence_after": "", "note": ""},
    "camera":           {"status": "consistent|lossy|altered|not_present", "severity": "none|minor|major", "evidence_before": "", "evidence_after": "", "note": ""},
    "lighting_color":   {"status": "consistent|lossy|altered|not_present", "severity": "none|minor|major", "evidence_before": "", "evidence_after": "", "note": ""},
    "scene":            {"status": "consistent|lossy|altered|not_present", "severity": "none|minor|major", "evidence_before": "", "evidence_after": "", "note": ""},
    "onscreen_text":    {"status": "consistent|lossy|altered|not_present", "severity": "none|minor|major", "evidence_before": "", "evidence_after": "", "note": ""}
  }
}
\end{promptbox}

\begin{promptbox}{Internal Consistency Output Format}
{
  "overall_verdict": "good|acceptable|bad",
  "internal_consistency_score": "xxx [score from 0 to 100]",
  "primary_issue": "subject_continuity|action_narrative_logic|temporal_state_continuity|spatial_consistency|physical_plausibility|audio_visual_sync|none",
  "summary": "",
  "dimensions": {
    "subject_continuity":         {"status": "consistent|ambiguous|inconsistent|not_present", "severity": "none|minor|major", "evidence": "", "note": ""},
    "action_narrative_logic":     {"status": "consistent|ambiguous|inconsistent|not_present", "severity": "none|minor|major", "evidence": "", "note": ""},
    "temporal_state_continuity":  {"status": "consistent|ambiguous|inconsistent|not_present", "severity": "none|minor|major", "evidence": "", "note": ""},
    "spatial_consistency":        {"status": "consistent|ambiguous|inconsistent|not_present", "severity": "none|minor|major", "evidence": "", "note": ""},
    "physical_plausibility":      {"status": "consistent|ambiguous|inconsistent|not_present", "severity": "none|minor|major", "evidence": "", "note": ""},
    "audio_visual_sync":          {"status": "consistent|ambiguous|inconsistent|not_present", "severity": "none|minor|major", "evidence": "", "note": ""}
  }
}
\end{promptbox}

\begin{promptbox}{Directorial Creation Output Format}
{
  "overall_verdict": "good|acceptable|bad",
  "directorial_creation_score": "xxx [score from 0 to 100]",
  "major_criterion_count": 0,
  "minor_criterion_count": 0,
  "primary_strength": "storyboard_timing|camera_design|editing_transition_design|lighting_color|story_richness_interest|core_prompt_centrality|sound_design|subject_performance_specificity|visual_style|none",
  "primary_issue": "storyboard_timing|camera_design|editing_transition_design|lighting_color|story_richness_interest|core_prompt_centrality|sound_design|subject_performance_specificity|visual_style|none",
  "summary": "",
  "dimensions": {
    "storyboard_timing": {
      "status": "complete|deficient",
      "severity": "none|minor|major",
      "missing_elements": [],
      "evidence": "",
      "note": ""
    },
    "camera_design": {
      "status": "complete|deficient",
      "severity": "none|minor|major",
      "missing_elements": [],
      "evidence": "",
      "note": ""
    },
    "editing_transition_design": {
      "status": "complete|deficient",
      "severity": "none|minor|major",
      "missing_elements": [],
      "evidence": "",
      "note": ""
    },
    "lighting_color": {
      "status": "complete|deficient",
      "severity": "none|minor|major",
      "missing_elements": [],
      "evidence": "",
      "note": ""
    },
    "story_richness_interest": {
      "status": "complete|deficient",
      "severity": "none|minor|major",
      "missing_elements": [],
      "evidence": "",
      "note": ""
    },
    "core_prompt_centrality": {
      "status": "complete|deficient",
      "severity": "none|minor|major",
      "missing_elements": [],
      "evidence": "",
      "note": ""
    },
    "sound_design": {
      "status": "complete|deficient",
      "severity": "none|minor|major",
      "missing_elements": [],
      "evidence": "",
      "note": ""
    },
    "subject_performance_specificity": {
      "status": "complete|deficient",
      "severity": "none|minor|major",
      "missing_elements": [],
      "evidence": "",
      "note": ""
    },
    "visual_style": {
      "status": "complete|deficient",
      "severity": "none|minor|major",
      "missing_elements": [],
      "evidence": "",
      "note": ""
    }
  }
}
\end{promptbox}

\subsection{Representative Case}
\label{app:all_task_cases}

We provide one representative case together with the outputs from all evaluated
models for direct comparison.

\begin{promptbox}{Input Prompt}
A sweet 9-year-old boy stands alone in a quiet neighborhood garden, looking slightly downcast. Moments later, his big brother walks up from behind with a kind smile and softly wraps him in a comforting embrace. The child(*@\textquoteright@*)s expression instantly brightens with joy. Warm afternoon light, vibrant animated style, fluid motion, touching and uplifting atmosphere, 3D family-film-style characters.

Duration: 15 seconds.
\end{promptbox}

\noindent\textbf{Model outputs.}

\begin{promptbox}{WanPE-397B}
The video consists of 4 shots. This 3D Animation, a work in the style of a Disney-Esque 3D Animation, is a short scene depicting a touching moment between two brothers. The overall art style is Cartoon/Animated, featuring characters with Exaggerated Semi-Realistic Proportions, rendered with Stylized PBR materials and illuminated by Stylized Realistic Lighting to create a warm, vibrant, and family-friendly atmosphere. The story shows a 9-year-old boy who is comforted by his older brother in a neighborhood garden. The scene is set in a quiet, sunlit neighborhood garden. A low, white picket fence lines the back of the garden, with lush green hedges and blooming flowers visible beyond it. A small, leafy tree stands on the left side of the scene. The ground is covered in soft, well-kept grass. The scene is bathed in the warm, golden light of a late afternoon. The video scene is set at late afternoon, the lighting overall led by soft Sunny Light, the image's light-dark contrast is medium-low, presenting a bright mid-tone characteristic. Color saturation is high, overlaid with a medium-intensity Pale Warm Yellow filter, the whole leaning toward a warm tone tendency, together creating a warm and uplifting emotional atmosphere. The narrative is presented from a third-person objective perspective using standard cinematic animation techniques with a measured editing pace, while the audio landscape combines a gentle, evolving piano score and subtle ambient sounds to create a touching and uplifting atmosphere.
Shot 1 [00.00-03.50s] The scene opens with a static, eye-level medium shot with a deep depth of field, employing a center composition. In the middle of the frame, a 9-year-old boy with a sweet face and large eyes stands alone on the soft, green grass. He wears a simple blue t-shirt and khaki shorts. The key light source is the soft, warm-yellow Sunny Light, forming a front side light effect on the boy and presenting a medium lighting ratio. The background features the white picket fence, lush green hedges, and blooming flowers, all softly lit by the afternoon sun. The boy performs an action of standing downcast. With his shoulders slightly slumped, he looks down at the ground, his expression downcast and lonely. His body remains still as a gentle, melancholic piano melody begins, accompanied by the faint, ambient sound of birds chirping and a soft breeze rustling leaves.
Shot 2 [03.50-06.50s] After a hard cut, the camera shifts to a medium long shot from a low angle, with a deep depth of field. The camera then performs a slow, stable dolly-in movement, moving forward to focus on the approaching character. The 9-year-old boy remains in the midground, his back partially to the camera. From the background, his big brother, a teenager with a kind, reassuring smile, wearing a gray hoodie and jeans, walks into the frame. The warm, golden sunlight creates a soft rim light on his form. The brother performs an action of walking towards the boy. He moves with a calm, steady gait, his soft footsteps audible on the grass. As he approaches, the piano melody continues, its tone becoming warmer and more hopeful.
Shot 3 [06.50-10.50s] A hard cut transitions to a static, over-the-shoulder medium close-up, with a shallow depth of field that focuses on the interaction. The 9-year-old boy's back is in the foreground, while the big brother is in the midground. The brother performs an action of embracing the boy. With a gentle, fluid motion, he wraps his arms around the boy from behind, the movement accompanied by the soft rustle of clothing. He rests his chin on the boy's shoulder, his expression full of warmth. The piano melody swells gently, becoming more prominent and emotional, as the ambient sounds of birds and breeze continue.
Shot 4 [10.50-15.00s] A final hard cut shifts to a close-up of the boy's face, captured at eye level with a shallow depth of field. The camera then performs a slow, stable zoom-in, tightening the frame on his expression. The boy is in the center of the frame, held in his brother's embrace. The boy's expression changes from downcast to joyful. Initially, his face is still, but then his eyes widen slightly, and a bright, genuine smile spreads across his face, his expression instantly brightening with joy. The camera slowly pushes in, emphasizing this emotional shift. The piano melody reaches a warm, uplifting crescendo, holding the note before beginning a slow fade, as the ambient sounds of the garden continue.
\end{promptbox}

\begin{promptbox}{WanPE-35B}
The video consists of 3 shots. This 3D family-film-style scene, in a warm, vibrant, and animated style, depicts a touching moment where a big brother comforts his downcast younger brother in a garden. The overall visual style is defined by a 3D Animation genre with a 3D family-film-style, featuring characters with a 3D family-film-style, rendered with a vibrant, polished, and cinematic quality. The scene is set in a quiet, well-kept neighborhood garden during the afternoon. The narrative is presented from a third-person objective perspective, using a standard cinematic animation style with a slow, deliberate editing pace that emphasizes the emotional connection. The audio landscape is composed of a gentle, warm piano score, ambient garden sounds, and a single line of soft, comforting dialogue, all contributing to a warm, uplifting, and touching atmosphere.
Shot 1 [00.00-04.00s] The scene opens with a medium shot at an eye-level angle, employing a center composition. The scene is set in a quiet, well-kept neighborhood garden during the afternoon, with a low brick wall and a wooden fence in the background, and a lush green lawn in the foreground. The video scene is set at day, the lighting overall led by soft Sunny Light, the image's light-dark contrast is medium-low, presenting a moderate mid-tone characteristic. Color saturation is high, overlaid with a medium intensity Pale Warm Yellow filter, the whole leaning toward a warm tone tendency, together creating a warm and uplifting emotional atmosphere. A 9-year-old boy, with a clean, innocent face, stands alone in the center of the frame, his body facing forward. He wears a simple t-shirt and shorts. The boy is looking slightly downcast. His shoulders are slightly slumped, and his expression is somber as he looks down at the ground, his posture conveying a sense of sadness. The camera performs a very slow, small-amplitude dolly-in, gradually tightening the frame on the boy, which enhances the feeling of his isolation. The key light source is the soft, warm-yellow afternoon sunlight, forming a front side light effect on the boy, presenting a medium lighting ratio. The background is softly lit by the same ambient daylight. The audio landscape is established with a gentle, warm piano melody, accompanied by the soft, ambient sound of a gentle breeze and distant birds chirping, creating a quiet, contemplative mood.
Shot 2 [04.00-09.00s] A hard cut transitions to a static, eye-level medium shot, maintaining the warm, sunny lighting. The 9-year-old boy remains in the center of the frame, still looking downcast. A 15-year-old big brother, with a kind, caring face, walks into the frame from the right, moving behind the boy. He wears a casual t-shirt and jeans. The big brother approaches with a gentle, kind smile, his movements fluid and deliberate. As he reaches the boy, he gently wraps his arms around the boy's shoulders, pulling him into a comforting embrace. The action is smooth and continuous, with the brother's arms moving in a gentle arc. The key light source is the soft, warm-yellow afternoon sunlight, forming a front side light effect on both boys, presenting a medium lighting ratio. The background is softly lit by the same ambient daylight. The piano melody continues, its warmth increasing slightly, and the sound of the brother's soft footsteps on the grass is heard, followed by the gentle rustle of clothing as he embraces the boy.
Shot 3 [09.00-15.00s] A hard cut shifts to a close-up of the 9-year-old boy's face, captured at an eye-level angle. The boy is being embraced by his brother, who is just out of frame. The boy's expression instantly brightens with joy. His eyes, previously downcast, lift to look up towards his brother, and a wide, genuine smile spreads across his face, his eyes crinkling with happiness. The camera performs a slow, small-amplitude dolly-in, emphasizing the boy's joyful reaction. The key light source is the soft, warm-yellow afternoon sunlight, forming a front side light effect on the boy, presenting a medium lighting ratio. The background is softly lit by the same ambient daylight. The piano melody swells to a warm, uplifting crescendo, and the ambient sound of a gentle breeze is present. The big brother speaks, his voice soft and comforting, "It's okay." The sound of a soft, contented sigh from the boy is heard, completing the emotional arc of the scene.
\end{promptbox}

\begin{promptbox}{H3-Context-IR}
integrated_multimodal_description: [Shot 1] 3D-animated, a medium shot with a slow push in on a 9-year-old boy standing alone in a stylized, vibrant neighborhood garden. The young boy features oversized, expressive hazel eyes, slightly messy chestnut-brown hair, and is dressed in a yellow-and-white striped cotton t-shirt tucked into blue denim overalls with bright red sneakers. He stands quietly on a winding cobblestone path, surrounded by oversized blooming blue hydrangeas and a pristine white picket fence in the background. Warm, golden afternoon sunlight filters through a dense canopy of green leaves above, casting distinct, dappled light rays across his downcast face. His shoulders slump forward as his gaze remains fixed on the ground, his right red sneaker lightly tapping against a small grey pebble. The camera pushes in steadily, emphasizing the boy's drooping posture and melancholic expression. [Shot 2] At 00:07.500, the camera cuts to a medium wide arc shot, circling the pair slowly as the boy's older brother steps into the frame from the background. The brother, a tall teenager with neatly parted brown hair, wears a thick forest-green hoodie and khaki trousers. He approaches the young boy from behind with a gentle, wide smile on his face. Upon reaching the child, the brother drops to one knee and softly wraps both arms around the boy's shoulders in a tight, comforting embrace. As the camera completes its arc to fully reveal the young boy's face, his expression instantly transforms. His large eyes widen, his eyebrows raise, and a massive, toothy smile breaks across his face. The golden afternoon light catches the edges of the brother's green hoodie and the boy's brown hair, creating a bright, glowing rim light around them as they hold the embrace.
overall_soundscape: A continuous ambient track of a gentle breeze and light bird chirps fills the background. In the foreground, a distinct, repetitive scrape and tap is clearly heard as the boy's sneaker hits the cobblestone and the pebble. This is followed by the audible, rhythmic thud of the older brother's footsteps approaching on the stone path, which transitions into a pronounced, loud rustle of heavy cotton fabric as the brother wraps his arms around the boy.
non_diegetic_music: Solo acoustic guitar accompanied by a gentle glockenspiel, slow tempo. The instrumentation begins with sparse, sustained notes, gradually building in volume and shifting to a bright, major-key chord progression in the second half, with no percussion.
\end{promptbox}

\begin{promptbox}{LTX-2.5-PE}
A medium shot frames a sweet 9-years-old boy with light brown hair and a pale blue t-shirt standing alone in a quiet, lush neighborhood garden filled with vibrant green foliage and colorful flowers, his gaze directed slightly downward, captured from a slightly low-angle perspective under warm afternoon light that casts soft shadows, the scene rendered in a vibrant animated style with fluid motion and rich, saturated film-grade colors, accompanied by a soft, gentle piano melody, a moment later, a tall, smiling big brother with dark hair and a casual light-colored shirt walks up from behind the boy, the camera smoothly tracks backward as the brother approaches, and he softly wraps his arms around the younger boy in a comforting embrace, the camera remains static as the embrace occurs, the boy's downcast expression instantly shifts to a bright, joyful smile, the warm cinematic lighting enhances the texture of the leaves and the soft glow on their faces, creating a touching and uplifting atmosphere with 3D family film-style characters and a subtle swell in the background music.
\end{promptbox}

\begin{promptbox}{LingBot-Video-PE}
{"comprehensive_description": {"scene_content_description": "In a lush, sun-drenched garden, a young boy with short brown hair and a white t-shirt stands alone, looking down with a somber expression. The background is filled with vibrant green foliage and a large, leafy tree. As the scene progresses, an older boy with curly brown hair, wearing a light blue t-shirt, enters from the right and approaches the younger boy. He gently places his hands on the younger boy's shoulders and then wraps his arms around him in a warm, comforting hug. The younger boy's expression shifts from sadness to a bright, joyful smile as he leans into the embrace. The lighting is warm and golden, suggesting late afternoon, creating a tender and emotional atmosphere.",
"camera_movement_description": "The camera remains essentially stationary throughout the video, maintaining a medium shot that focuses on the two boys. There is a very slight, natural handheld tremor, but no intentional panning, tilting, or zooming occurs."},
"camera_info": {"color": "Warm", "frame_size": "Medium", "shot_type_angle": "Low angle", "lens_size": "Medium", "composition": "Center", "lighting": "Hard light", "lighting_type": "Daylight"},
"world_knowledge": [],
"prominent_elements": [{"name": "young boy", "description": "A young boy with short brown hair and a white t-shirt, appearing sad at the start of the video.", "actions": [{"timestamp": "[0.0s - 4.5s]", "action": "stands still, looking down with a sad expression"}, {"timestamp": "[4.5s - 10.5s]", "action": "remains still as the older boy approaches and hugs him"}, {"timestamp": "[10.5s - 13.0s]", "action": "smiles broadly and leans into the hug"}, {"timestamp": "[13.0s - 15.0s]", "action": "stands still, smiling while being hugged"}], "location": "center of the frame", "relative_size": "medium", "shape_and_color": "humanoid shape, wearing a white t-shirt", "texture": "soft skin, cotton t-shirt", "appearance_details": "short brown hair, white t-shirt with a small graphic on the chest", "relationship": "being comforted by the older boy", "orientation": "facing forward, slightly tilted down", "pose": "standing upright, then leaning into a hug", "expression": "changes from sad and downcast to happy and smiling", "clothing": "white t-shirt", "gender": "male", "skin_tone_and_texture": "fair, smooth skin"},
{"name": "older boy", "description": "An older boy with curly brown hair and a light blue t-shirt, acting as a comforting figure.", "actions": [{"timestamp": "[4.5s - 7.0s]", "action": "enters from the right and walks toward the younger boy"}, {"timestamp": "[7.0s - 10.5s]", "action": "places hands on the younger boy's shoulders and hugs him"}, {"timestamp": "[10.5s - 15.0s]", "action": "holds the younger boy in a hug, smiling"}], "location": "enters from the right, moves to the center", "relative_size": "medium", "shape_and_color": "humanoid shape, wearing a light blue t-shirt", "texture": "curly hair, cotton t-shirt", "appearance_details": "curly brown hair, light blue t-shirt", "relationship": "comforting the younger boy", "orientation": "facing left toward the younger boy", "pose": "walking, then standing and hugging", "expression": "kind and smiling", "clothing": "light blue t-shirt", "gender": "male", "skin_tone_and_texture": "fair, smooth skin"}]}
\end{promptbox}

\begin{promptbox}{SCMaPR}
A sweet 9-year-old boy stands alone in a quiet neighborhood garden, looking slightly downcast with a slightly downcast expression under warm afternoon light. His older brother walks up gently from behind and softly wraps his arms around the younger boy in a comforting embrace. The younger boy(*@\textquoteright@*)s face immediately brightens with joy in direct response to this tender physical contact. The scene is rendered in a vibrant animated style with fluid motion and 3D family-film-style characters to create a touching and uplifting atmosphere.
\end{promptbox}

\begin{promptbox}{VPO}
A sweet 9-year-old boy with tousled hair and a plain T-shirt stands alone in a quiet neighborhood garden, surrounded by blooming flowers and a wooden bench under the warm sunlight. His posture is slightly slouched, and his gaze is cast downward, suggesting a moment of quiet melancholy. The scene captures the soft rustle of leaves and the gentle sway of nearby trees under the golden afternoon light, adding to the serene yet slightly somber mood. Moments later, his big brother, slightly taller and wearing a casual hoodie, walks up from behind with a warm and kind smile. The boy(*@\textquoteright@*)s expression softens as his older brother approaches, and with a heartfelt gesture, he gently wraps his arms around him in a comforting embrace. The younger boy(*@\textquoteright@*)s face lights up with pure joy, radiating warmth and the unspoken bond of sibling love. The animated style features soft, fluid motion and vibrant colors, enhancing the heartfelt and uplifting atmosphere of the moment. The background includes subtle details like a colorful butterfly fluttering near the flowers and distant chirping birds, reinforcing the wholesome, family-film-style aesthetic.
\end{promptbox}

\begin{promptbox}{Prompt-A-Video-Cog}
In a serene 3D neighborhood garden, a cherubic 9-year-old boy with a softly tousled hairdonned a lightweight jacket, standing alone under the embracing branches of a oak tree. His eyes, cast downward, whisper stories of a gentle sorrow. As his stalwart big brother, with a warm smile, wraps him in a comforting embrace, the younger boy's expression transforms, brightening with joy. The brothers enclave is bathed in the warm, honeyed light of a late afternoon sun. The animation is vibrant, with fluid motions, captivating in a style reminiscent of timeless family films. The scene is a powerful testament to the embracing love of a big brother, as the animation exudes a touching, uplifting atmosphere that lasts for 15 captivating seconds.
\end{promptbox}

\begin{promptbox}{Prompt-A-Video-OS}
In a vibrant neighborhood garden bathed in warm afternoon light, a 9-year-old boy with a downcast expression sits alone. His big brother, with a kind face, walks into the frame and envelops him in a comforting hug. The boy's demeanor instantly brightens with joy. The video is rendered in a 3D animated style, characteristic of family films, and exudes a touching and uplifting atmosphere. (15 seconds)
\end{promptbox}

\begin{promptbox}{RAPO}
a young boy wearing a red shirt and sitting on the ground. He is then seen being hugged by his big brother. The scene is set in a warm and sunny afternoon. The animation is colorful and vibrant. The boy's expression changes from sadness to joy as he is being hugged by his big brother.
\end{promptbox}

\section{Additional Experimental Details}
\label{app:additional_experiments}

\subsection{Implementation Details}
\label{app:implementation_details}
We evaluate the ten prompt-enhancement systems listed in
Table~\ref{tab:main_results}, together with the original prompt as a
no-enhancement baseline. All systems are evaluated on T2V-PE, while systems
supporting visual inputs are additionally evaluated on I2V-PE and R2V-PE.
Proprietary systems are accessed through their official interfaces, and
released models are run using their recommended inference settings.

For each case, modality-aware evidence is extracted once and cached for all
methods. GPT-5.4~\citep{gpt54} extracts facts for image-only inputs, and
Qwen3.5-Omni-Plus processes all assets for cases containing video references.
GPT-5.4 then evaluates every enhanced prompt using the same criterion-specific
rubrics. We report criterion pass rates and aggregate Semantic Fidelity,
Internal Consistency, and Directorial Creation scores.
The additional experiments analyze temporal complexity, reference complexity,
and output length. Human studies further test prompt-level alignment and
downstream utility using videos generated by Wan3.0 and MiniMax-H3.
For downstream video generation, we access Wan3.0 through its online API with
prompt enhancement disabled, while MiniMax-H3 is run locally on eight NVIDIA
A100 GPUs.

\subsection{Length-Stratified Analysis}
\label{app:length_analysis}

Prompt enhancement often transforms concise inputs into substantially longer
descriptions. To test whether output length alone yields better descriptive
adequacy, we analyze H3-Context-IR separately on each task. Valid outputs are
sorted by token length and divided into four equal-sized quartiles, from the
shortest (\(L1\)) to the longest (\(L4\)).

\begin{table}[h!]
\centering
\small
\caption{\textbf{Length-stratified results for H3-Context-IR.} Quartiles are
defined independently within each task. DC denotes Directorial Creation.}
\label{tab:length_analysis}
\setlength{\tabcolsep}{15pt}
\renewcommand{\arraystretch}{1}
\begin{tabular}{llrrrr}
\toprule
Task & Metric & \(L1\) & \(L2\) & \(L3\) & \(L4\) \\
\midrule
T2V-PE & Avg.\ tokens & 394.7 & 490.9 & 584.9 & 861.1 \\
       & DC           & 79.32 & 78.60 & 77.35 & 74.13 \\
\midrule
I2V-PE & Avg.\ tokens & 819.1 & 945.7 & 1066.6 & 1382.0 \\
       & DC           & 75.00 & 75.20 & 75.99 & 67.20 \\
\midrule
R2V-PE & Avg.\ tokens & 809.9 & 1015.7 & 1164.4 & 1479.3 \\
       & DC           & 79.37 & 73.75 & 72.07 & 69.28 \\
\bottomrule
\end{tabular}
\end{table}

Across all three tasks, the longest quartile underperforms the shortest by
5.19 to 10.09 DC points. Thus, longer outputs do not consistently provide
better directorial guidance. This result is consistent with the evaluation
design: DC rewards concrete, relevant, and executable specifications for
temporal planning, cinematography, narrative progression, subject
performance, and sound, rather than verbosity or repeated template language.

\section{Human Expert Annotation}
\label{app:human_evaluation}

\subsection{Pairwise Annotation Protocol}

The human study evaluates alignment at two levels: (i) the quality of enhanced
prompts and (ii) the quality of downstream videos generated from those
prompts. We recruit 60 experts and provide task-specific training before
annotation. Each expert evaluates approximately 50 T2V-PE pairs, 20 I2V-PE
pairs, and 10 R2V-PE pairs. The comparison pool contains 11 systems for
T2V-PE, 4 for I2V-PE, and 3 for R2V-PE, including the original prompt as the
no-enhancement baseline in every task. 

\noindent\textbf{Pairwise judgment procedure.}

For enhanced-prompt evaluation, annotators inspect the original instruction,
target duration, reference assets when applicable, and an anonymized pair of
enhanced prompts. Annotators make a separate choice for Semantic Fidelity,
Internal Consistency, and Directorial Creation. For each dimension, they
select the better candidate, both good, or both bad.
For downstream-video evaluation, annotators inspect the source request and
references together with an anonymized pair of generated videos and make the
same dimension-specific judgment. Candidate identities are hidden, and the
left and right positions are randomized independently for every comparison.

We aggregate the pairwise choices with a Bradley-Terry model to obtain a
system ranking for each task, evaluation level, and dimension. We then compute
the Spearman correlation between this expert-derived ranking and the
corresponding automatic dimension scores. The resulting coefficients are
reported in Table~\ref{tab:human_downstream_alignment}.

\noindent\textbf{Quality control and agreement.}

All experts complete training before formal annotation. We test each
annotator's selections for a significant left or right position preference.
Annotators exhibiting significant position bias are excluded, together with
all of their judgments, before fitting the Bradley--Terry models and computing
the correlations.

\subsection{Annotation Interface}

The interface presents the original user prompt and target duration at the
top, followed by reference assets for I2V-PE and R2V-PE. Prompt-level
evaluation displays two anonymized enhanced prompts, while video-level
evaluation replaces them with two generated videos. The three evaluation
dimensions are shown as independent rating rows.

\begin{figure*}[p]
\centering
\textbf{T2V-PE Annotation Interface}\\[1mm]
\includegraphics[width=0.80\textwidth,height=0.40\textheight,keepaspectratio]{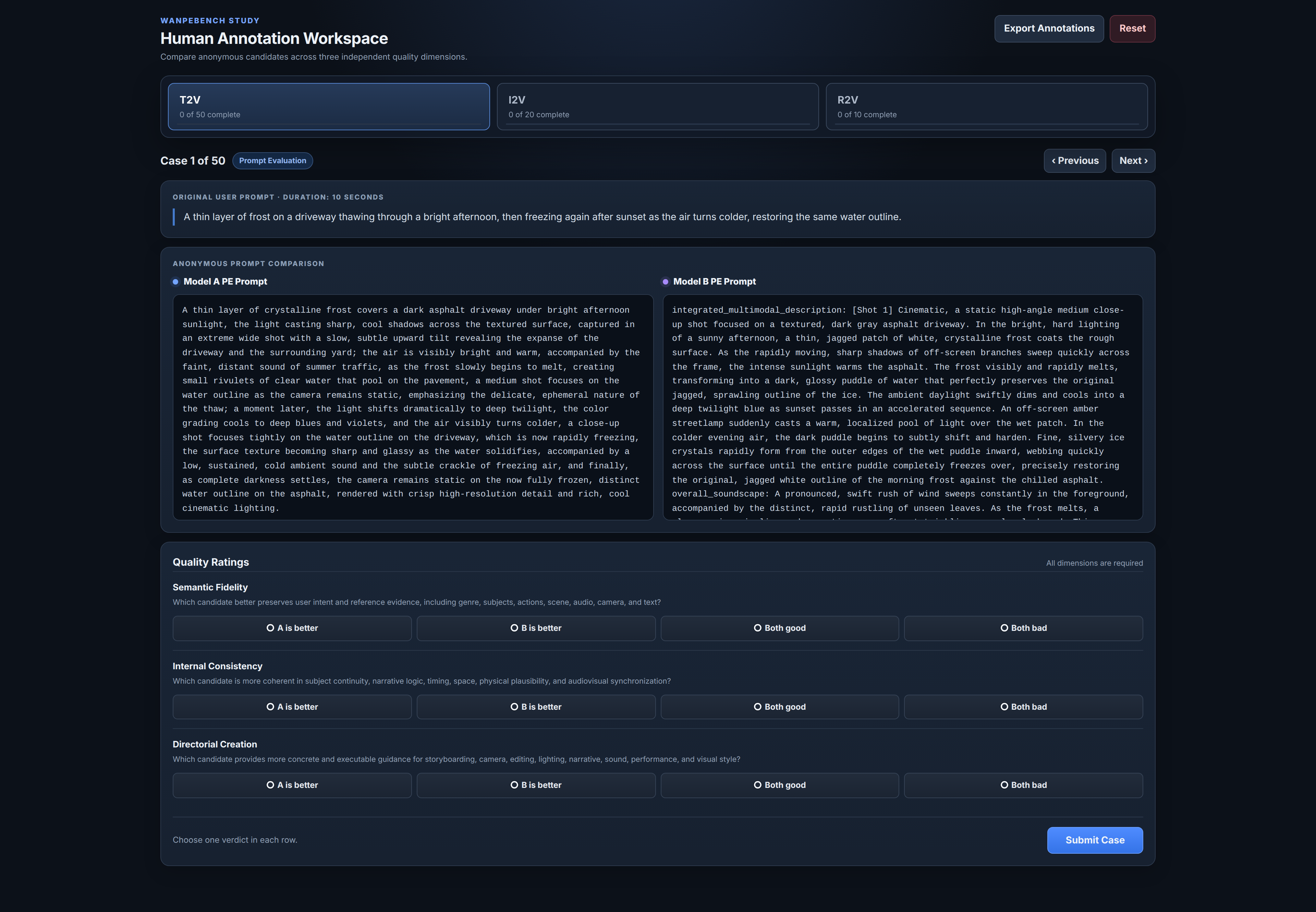}\\[-1mm]
\small (a) Prompt-level evaluation
\vspace{2mm}

\includegraphics[width=0.80\textwidth,height=0.40\textheight,keepaspectratio]{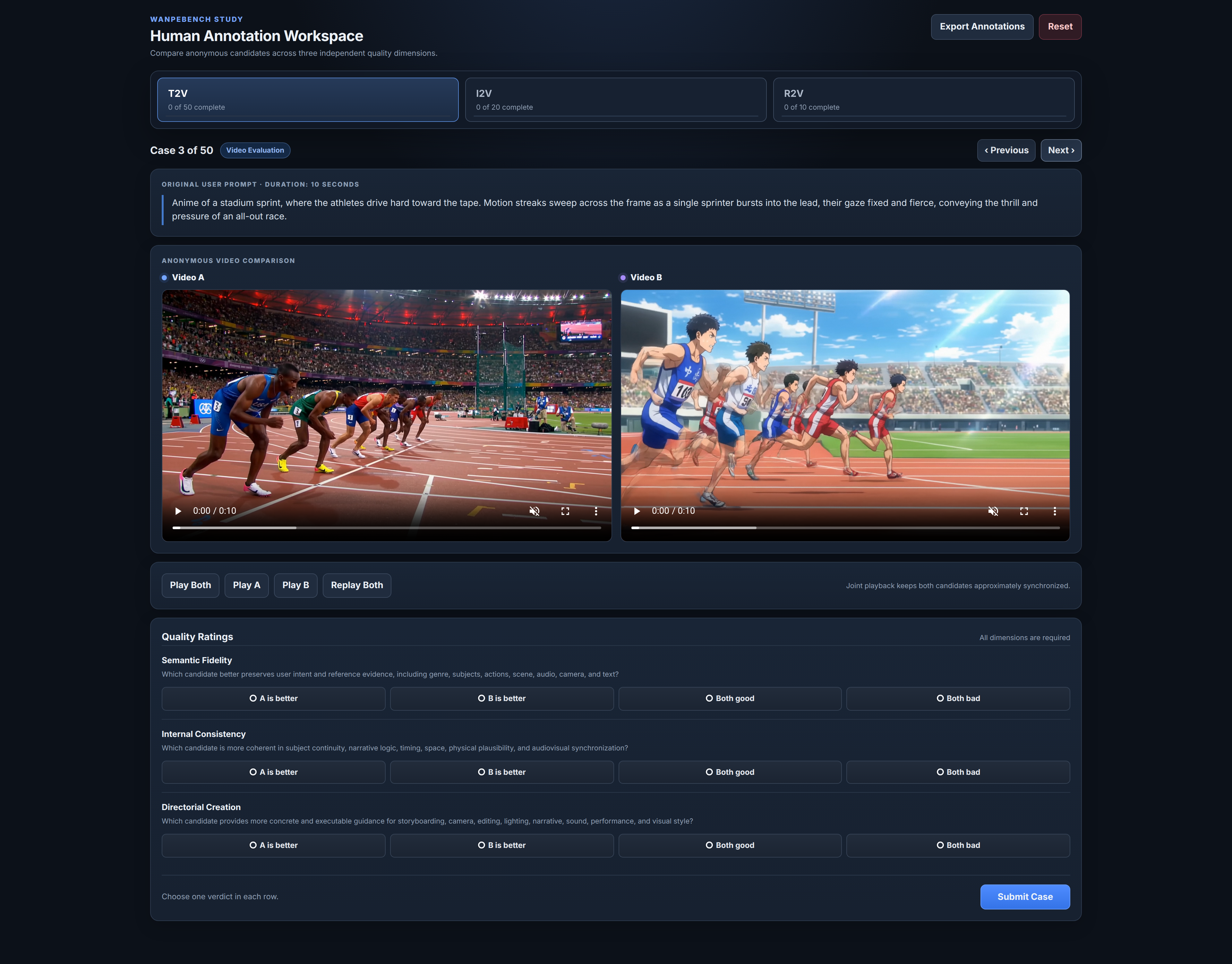}\\[-1mm]
\small (b) Video-level evaluation
\caption{\textbf{T2V-PE annotation interface.} Anonymous prompts and generated
videos are compared against the text instruction and target duration.}
\label{fig:human_evaluation_interface_t2v}
\end{figure*}

\begin{figure*}[p]
\centering
\textbf{I2V-PE Annotation Interface}\\[1mm]
\includegraphics[width=0.80\textwidth,height=0.40\textheight,keepaspectratio]{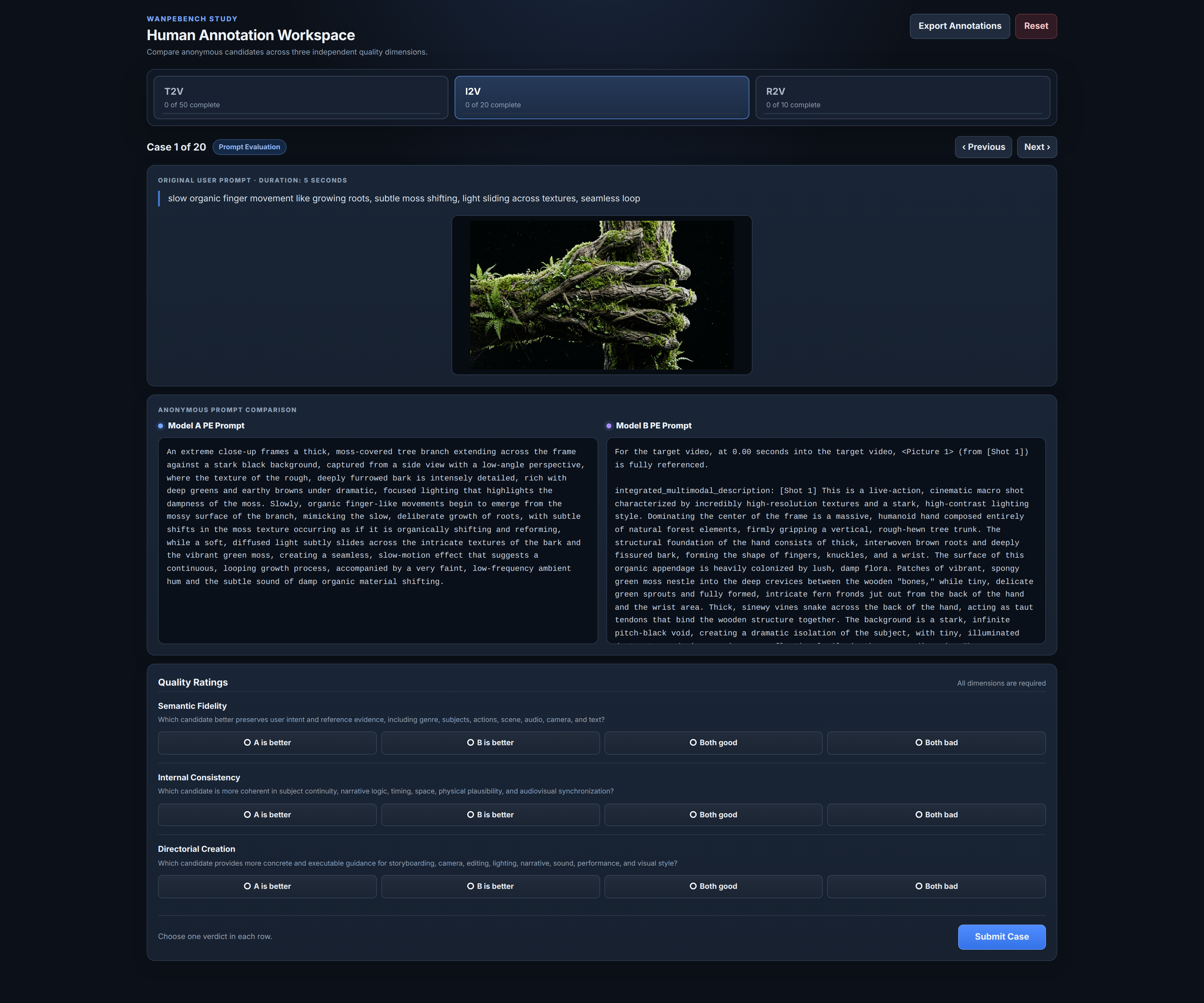}\\[-1mm]
\small (a) Prompt-level evaluation
\vspace{2mm}

\includegraphics[width=0.80\textwidth,height=0.40\textheight,keepaspectratio]{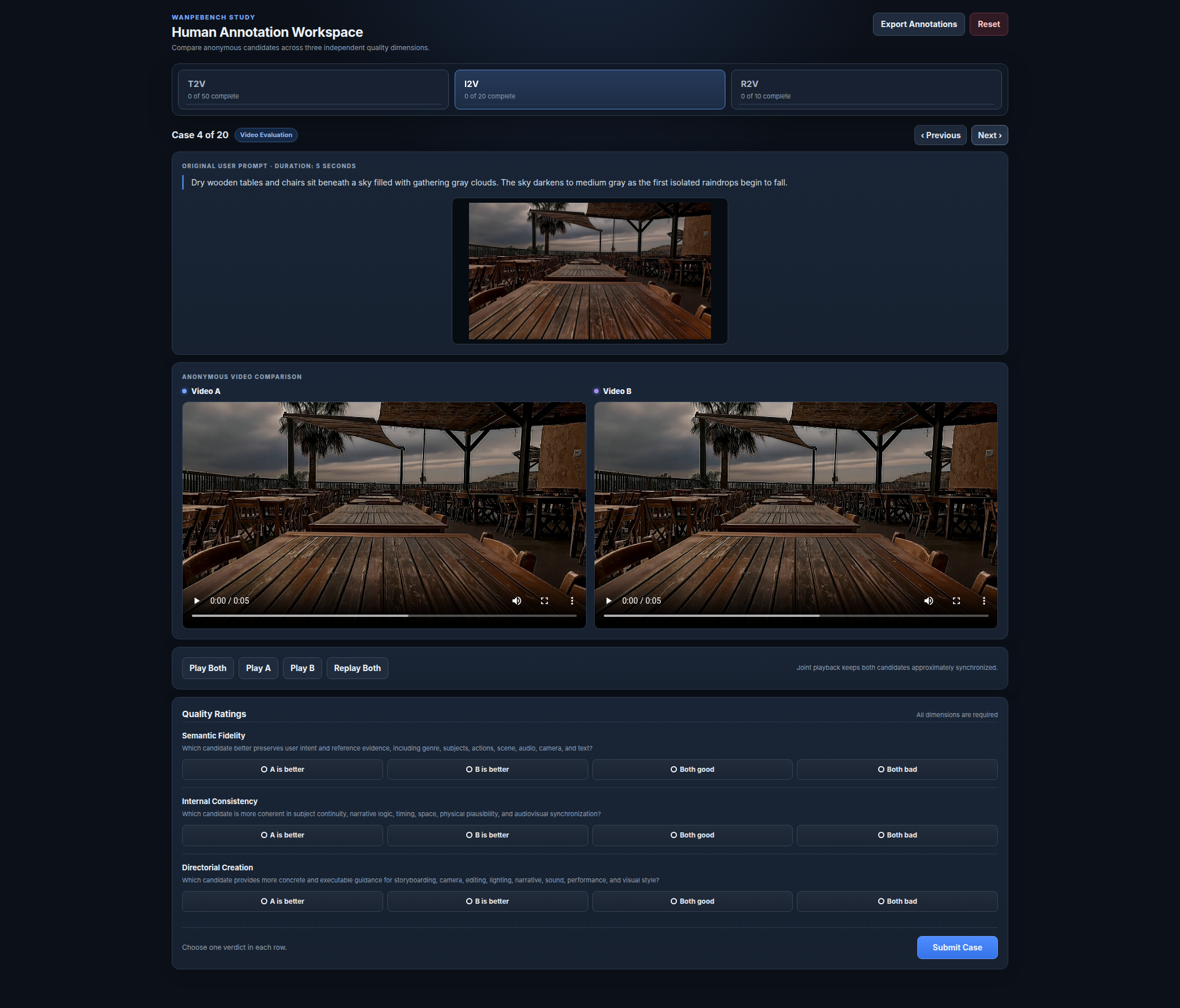}\\[-1mm]
\small (b) Video-level evaluation
\caption{\textbf{I2V-PE annotation interface.} The initial-frame image is
displayed with the instruction for prompt-level and video-level comparison.}
\label{fig:human_evaluation_interface_i2v}
\end{figure*}

\begin{figure*}[p]
\centering
\textbf{R2V-PE Annotation Interface}\\[1mm]
\includegraphics[width=0.80\textwidth,height=0.40\textheight,keepaspectratio]{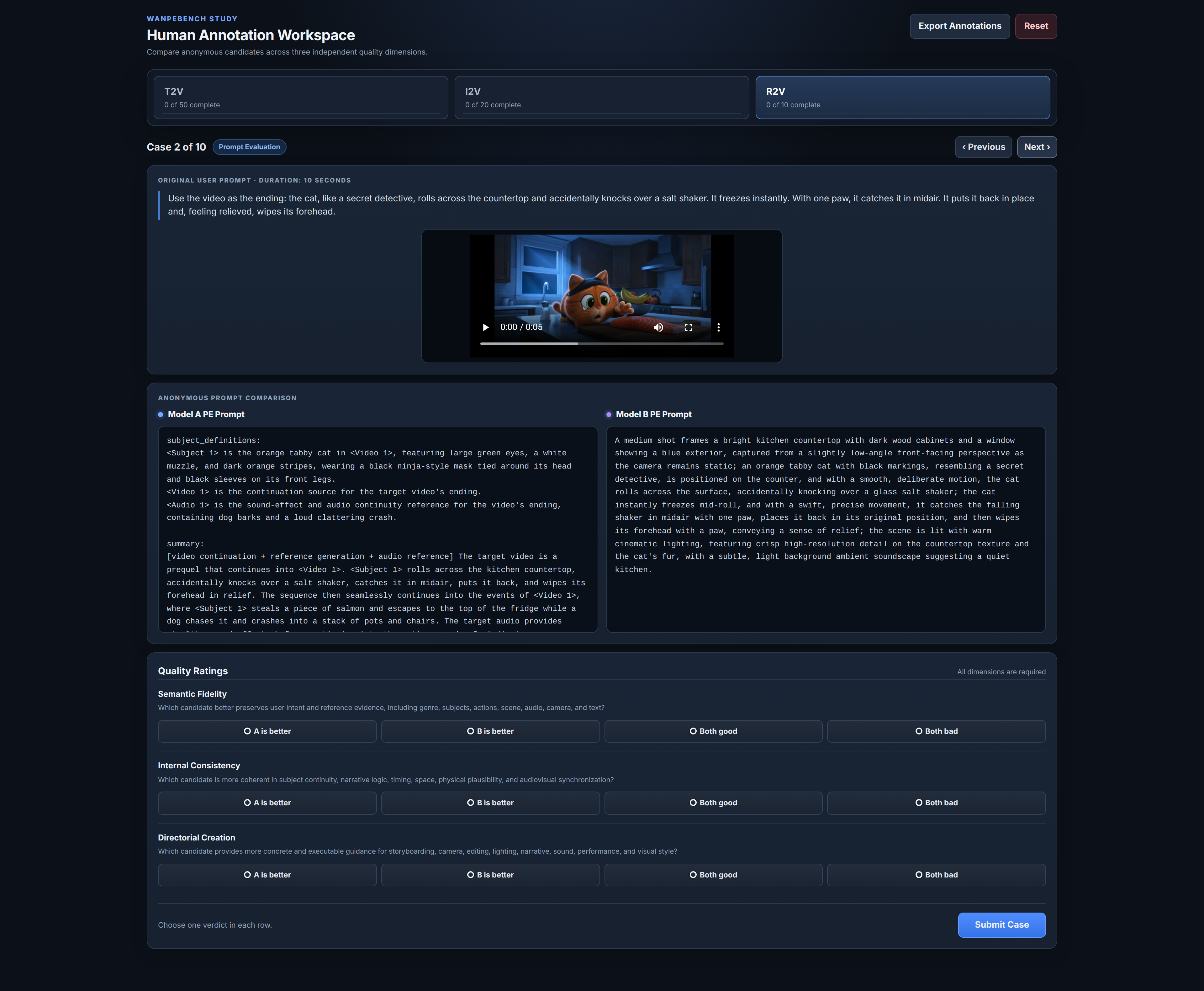}\\[-1mm]
\small (a) Prompt-level evaluation
\vspace{2mm}

\includegraphics[width=0.80\textwidth,height=0.40\textheight,keepaspectratio]{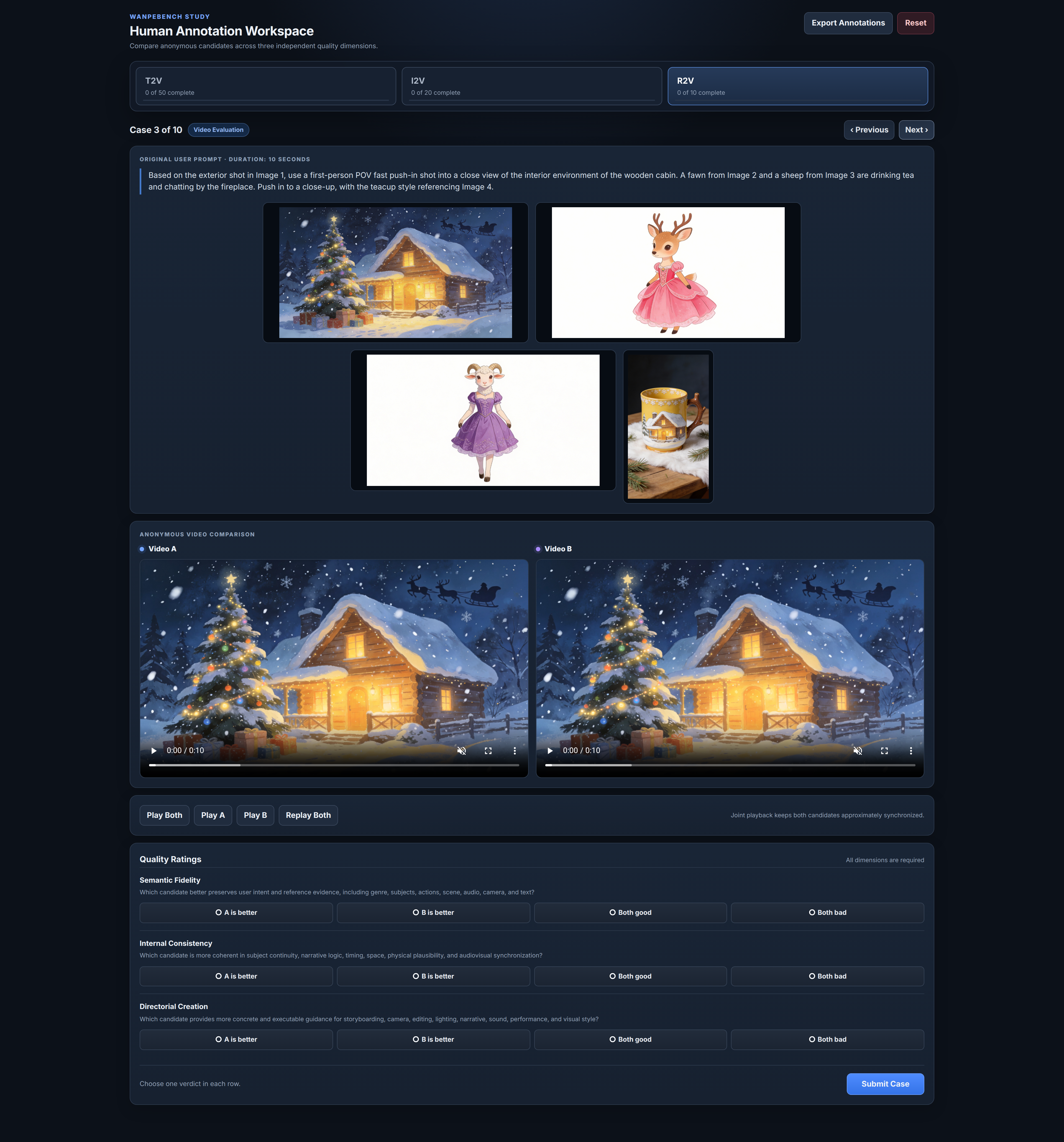}\\[-1mm]
\small (b) Video-level evaluation
\caption{\textbf{R2V-PE annotation interface.} All reference images and
videos are displayed to support cross-asset grounding judgments.}
\label{fig:human_evaluation_interface_r2v}
\end{figure*}

\noindent\textbf{Interface behavior.}
Candidate identities are replaced by A and B, and their left and right
positions are randomized independently for each case. Video-level evaluation
supports synchronized playback, individual playback, and replay. The
interface requires one response for each dimension before submission and
allows annotators to revise their selections before completing the case. Figures~\ref{fig:human_evaluation_interface_t2v}--\ref{fig:human_evaluation_interface_r2v} show the annotation interfaces for the three PE settings.

\section{Ethics, Privacy, and Licensing}
\label{app:ethics_privacy_licensing}

\bench{} prompts are generated from an expert-designed taxonomy and undergo
manual verification. During curation, we remove personal information,
identifiable private individuals, sensitive political content, and material
that may raise privacy or cultural concerns. Reference assets originate from
public benchmarks or model-assisted generation. We retain source and license
metadata and release only assets permitted for redistribution, providing
source identifiers for restricted material. Videos generated for downstream
validation are used only for evaluation and are not included in the benchmark
release. We release the verified prompts, legally shareable reference assets,
and evaluation framework to support reproducible research.

\section{Limitations}
\label{app:limitations}
One limitation is the scale of the benchmark. Although 1,100
expert-verified cases provide broad coverage across three PE settings, 35 task
types, and diverse conditioning patterns, they cannot represent the full
range of rapidly evolving video-generation requests. Future work will expand
\bench{} with additional domains, languages, reference configurations, and
emerging control formats, enabling more comprehensive evaluation and
supporting the development of increasingly general prompt enhancers.

A second limitation is the comparatively limited model coverage for I2V-PE
and R2V-PE. Unlike T2V-PE, few existing methods support multimodal prompt
enhancement, and only a small number of proprietary systems, such as
H3-Context-IR, provide accessible PE interfaces. Consequently, our I2V-PE and
R2V-PE evaluations include fewer baselines. We will continue updating
\bench{} as more multimodal PE methods become available, enabling broader
comparisons across open and proprietary systems.